\documentclass{article} %
\usepackage{iclr2026_conference,times}

\usepackage[table,dvipsnames]{xcolor}
\usepackage{tikz}
\usepackage{amsmath}

\usepackage[utf8]{inputenc} %
\usepackage[T1]{fontenc}    %
\usepackage{amssymb}
\usepackage{hyperref}       %
\usepackage{url}            %
\usepackage{amsfonts}       %
\usepackage{nicefrac}       %
\usepackage{microtype}      %
\usepackage{graphicx}       %
\usepackage{subcaption}
\usepackage{pgfplots}
\usepackage{algorithm}
\usepackage{algorithmic}
\usepackage{wrapfig}
\usepackage{pifont}
\usepackage[font=small,labelfont=bf]{caption}
\usepackage{nicematrix}
\usepackage{booktabs}       %
\usepackage{enumitem}
\usepackage{amsthm}
\usepackage{mathtools}
\usepackage{listings}
\usepackage{cleveref}

\usetikzlibrary{matrix,positioning,fit,calc,patterns}
\pgfplotsset{compat=1.17}

\newcolumntype{L}[1]{>{\raggedright\arraybackslash}p{#1}}
\newcolumntype{C}[1]{>{\centering\arraybackslash}p{#1}}

\definecolor{level0}{HTML}{403d39}  %
\definecolor{level1}{HTML}{252422}   %
\definecolor{level2}{HTML}{84a98c}    %
\definecolor{level3}{HTML}{E07A5F}    %
\definecolor{level4}{HTML}{3D405B}    %
\newcommand{\lambdalevel}[2]{%
  \ifnum#1=0
    {\textcolor{level0}{\lambda^{(#1)}_{#2}}}%
  \else\ifnum#1=1
    {\textcolor{level1}{\lambda^{(#1)}_{#2}}}%
  \else\ifnum#1=2
    {\textcolor{level2}{\lambda^{(#1)}_{#2}}}%
  \else\ifnum#1=3
    {\textcolor{level3}{\lambda^{(#1)}_{#2}}}%
  \else\ifnum#1=4
    {\textcolor{level4}{\lambda^{(#1)}_{#2}}}%
  \else
    {\Lambda^{(#1)}_{#2}}%
  \fi\fi\fi\fi\fi
}

\makeatletter
\def\adl@drawiv#1#2#3{%
        \hskip.5\tabcolsep
        \xleaders#3{#2.5\@tempdimb #1{1}#2.5\@tempdimb}%
                #2\z@ plus1fil minus1fil\relax
        \hskip.5\tabcolsep}
\newcommand{\cdashlinelr}[1]{%
  \noalign{\vskip\aboverulesep
           \global\let\@dashdrawstore\adl@draw
           \global\let\adl@draw\adl@drawiv}
  \cdashline{#1}
  \noalign{\global\let\adl@draw\@dashdrawstore
           \vskip\belowrulesep}}
\makeatother

\definecolor{codegreen}{rgb}{0,0.6,0}
\definecolor{codegray}{rgb}{0.5,0.5,0.5}
\definecolor{codepurple}{rgb}{0.07,0,0.53}
\definecolor{codered}{RGB}{189,41,0}
\definecolor{codecomment}{RGB}{153,153,153}
\definecolor{backcolour}{rgb}{0.96,0.96,0.96}
\definecolor{mygreen}{rgb}{0.0, 0.5, 0.0}
\definecolor{royalblue}{rgb}{0.0, 0.14, 0.4}
\definecolor{egyptianblue}{rgb}{0.06, 0.2, 0.65}
\definecolor{royalazure}{rgb}{0.0, 0.22, 0.66}
\definecolor{portlandorange}{rgb}{1.0, 0.35, 0.21}
\definecolor{saddlebrown}{RGB}{139,69,19}
\definecolor{sienna}{RGB}{183,105,68}
\definecolor{saddlebrown}{RGB}{139,69,19}

\lstdefinestyle{mystyle}{
    backgroundcolor=\color{backcolour},   
    commentstyle=\color{codegreen},
    keywordstyle=\color{codered},
    numberstyle=\tiny\color{codegray},
    stringstyle=\color{codepurple},
    emphstyle=\color{codered},    
    basicstyle=\ttfamily\tiny,
    breakatwhitespace=false,         
    breaklines=true,                 
    captionpos=b,                    
    keepspaces=true,                 
    numbers=left,                    
    numbersep=5pt,                  
    showspaces=false,                
    showstringspaces=false,
    showtabs=false,   
    morekeywords={>,<,.,;,-,!,=,~},
    tabsize=2
}
\usepackage{amsmath,amsfonts,bm}

\def\eqref#1{equation~\ref{#1}}

\def\1{\bm{1}}

\def\rmS{{\mathbf{S}}}

\def\vh{{\bm{h}}}

\def\vk{{\bm{k}}}

\def\vo{{\bm{o}}}

\def\vq{{\bm{q}}}

\def\vv{{\bm{v}}}

\def\vx{{\bm{x}}}

\def\mA{{\mathbf{A}}}

\def\mC{{\mathbf{C}}}
\def\mD{{\mathbf{D}}}

\def\mI{{\mathbf{I}}}

\def\mK{{\mathbf{K}}}
\def\mL{{\mathbf{L}}}
\def\mM{{\mathbf{M}}}

\def\mO{{\mathbf{O}}}
\def\mP{{\mathbf{P}}}
\def\mQ{{\mathbf{Q}}}
\def\mR{{\mathbf{R}}}
\def\mS{{\mathbf{S}}}
\def\mT{{\mathbf{T}}}
\def\mU{{\mathbf{U}}}
\def\mV{{\mathbf{V}}}

\def\mY{{\mathbf{Y}}}

\def\mLambda{{\mathbf{\Lambda}}}

\def\mSigma{{\mathbf{\Sigma}}}

\DeclareMathAlphabet{\mathsfit}{\encodingdefault}{\sfdefault}{m}{sl}
\SetMathAlphabet{\mathsfit}{bold}{\encodingdefault}{\sfdefault}{bx}{n}
\newcommand{\tens}[1]{\bm{\mathsfit{#1}}}

\def\tM{{\tens{M}}}

\def\gO{{\mathcal{O}}}

\def\sR{{\mathbb{R}}}

\title{Log-Linear Attention}

\author{
\textbf{Han Guo}$^{1}$\thanks{Equal contribution.} \quad
\textbf{Songlin Yang}$^{1}$\footnotemark[1] \quad
\textbf{Tarushii Goel}$^{1}$ \quad
\textbf{Eric P. Xing}$^{3}$ \quad
\textbf{Tri Dao}$^{2}$ \quad
\textbf{Yoon Kim}$^{1}$ \vspace{2mm} \\
$^{1}$Massachusetts Institute of Technology \quad $^{2}$Princeton University, Together AI \vspace{1mm} \\
$^{3}$Carnegie Mellon University, Mohamed bin Zayed University of AI, GenBio AI  
\vspace{2mm} \\ 
\texttt{hanguo@mit.edu}
}

\iclrfinalcopy %
\begin{document}

\maketitle

\vspace{-4mm}
\begin{abstract}
\vspace{-2mm}
The attention mechanism in Transformers is an important primitive for accurate and scalable sequence modeling. Its quadratic-compute and linear-memory complexity however remain significant bottlenecks. Linear attention and state-space models enable linear-time, constant-memory sequence modeling and can moreover be trained efficiently through matmul-rich parallelization across sequence length. However, at their core these models are still RNNs, and thus their use of a fixed-size hidden state to model the context is a fundamental limitation. This paper develops log-linear attention, an attention mechanism that balances linear attention's efficiency and the expressiveness of softmax attention. Log-linear attention replaces the fixed-size hidden state with a logarithmically growing set of hidden states. We show that with a particular growth function, log-linear attention admits a similarly matmul-rich parallel form whose compute cost is log-linear in sequence length. Log-linear attention is a general framework and can be applied on top of existing linear attention variants. As case studies, we instantiate log-linear variants of two recent architectures---Mamba-2 and Gated DeltaNet---and find they perform well compared to their linear-time variants.\footnote{Code available at \url{https://github.com/HanGuo97/log-linear-attention}.}
\end{abstract}

\section{Introduction}

\label{sec:intro}

The attention layer \citep{bahdanau2014neural} is a core building block of modern deep learning architectures, most notably in the Transformer architecture \citep{vaswani2017attention}. For training, attention can be parallelized across sequence length through reformulating the computation as a series of matrix-matrix multiplications (matmuls), which can enable efficient training on modern  accelerators such as GPUs and TPUs. However, the compute cost of attention  grows quadratically and its memory cost grows linearly with respect to sequence length;  despite the wallclock efficiency improvements obtained from hardware-optimized  implementations \citep{dao2022flashattention,dao2023flashattention2,shah2024flashattention,liu2023ring,kwon2023efficient}, this quadratic-compute linear-memory cost  is a fundamental limitation in enabling new applications and serves as a significant bottleneck in existing ones.

Linear attention \citep{katharopoulos2020transformers} replaces the softmax kernel with a simple linear kernel (i.e., dot product)  to derive the ``attention'' scores. The use of a linear  kernel makes it possible to reformulate  linear attention  as a linear RNN  with matrix-valued hidden states, and thus  linear attention enables linear-time, constant-memory sequence modeling.\footnote{Thus there are three senses in which linear attention is \emph{linear}: the use of a linear kernel, its reformulation as a linear RNN where the hidden state is a linear function of the previous  state, and its linear-time complexity.} For training, linear attention can be parallelized across sequence length via a chunking mechanism where a sequence is split up into chunks and the computations across chunks are performed in parallel \citep{hua2022transformer,sun2023retentive,yang2024parallelizing,dao2024transformers}. The complexity of this chunkwise parallel algorithm is subquadratic in sequence  length but still rich in matmuls,\footnote{Unlike parallel scan \citep{blelloch1990prefix} which can also parallelize linear attention across sequence length but consists mostly of  elementwise operations instead of matmuls.} leading to hardware-efficient implementations \citep{yang2024fla,qin2024lightning,beck2025tiled} that obtain practical wallclock improvements over optimized implementations of softmax attention.
While early versions of linear attention generally underperformed softmax attention \citep{kasai2021finetuning,peng_random_2021,mao_fine-tuning_2022,qin2022cosformer,sun2023retentive}, modern variants with data-dependent multiplicative gates \citep{yang2024parallelizing,qin_hgrn2_2024,peng2024eagle}---which include state-space models (SSMs) such as Mamba \citep{Gu2023MambaLS,dao2024transformers}---and delta-rule-based structured transition matrices \citep{schlag_linear_2021,yang2024parallelizing,yang2024gated,grazzi2025unlocking,siems2025deltaproduct,peng2025rwkv} have led to significant improvements. However, despite these improvements linear attention's use of a fixed-sized hidden state is a fundamental  limitations when it comes to certain capabilities such as associative recall over a given context \citep{arora2024simple}. 

This paper develops log-linear attention as a middle ground between linear attention and full softmax
attention. Instead of using a single hidden state matrix to represent the history (as in linear attention/SSMs), log-linear attention maintains a growing set of hidden states where the set size grows
logarithmically with respect to sequence length. With a particular choice of the growth function,
we show that log-linear attention admits a matmul-rich “parallel form” for training which involves
replacing the lower-triangular causal mask in ordinary linear attention with a data-dependent hierarchical matrix, which enables subquadratic training; in particular we show that the compute cost
of log-linear attention is log-linear in sequence length (hence the name), while its memory cost is
logarithmic. Log-linear attention is a general framework for sequence modeling and can be used to
generalize existing linear attention models. As case studies, we use the framework on two popular
recent models, Mamba-2 \citep{dao2024transformers} and Gated DeltaNet \citep{yang2024gated}, to derive log-linear variants of both models,
and find that these variants perform well compared to their original linear variants.

\section{Background: A Structured Matrix View of Efficient Attention}
\label{sec:efficient_attention_with_structured_matrices}

\begin{table}[!t]
\centering
\scriptsize
\begin{tabular}{r|lllll}
\textbf{Model} & \textbf{$\mA$} & \textbf{$\mM$} \textbf{(Data Dependent?)} & \textbf{Training Algorithm / Time} & \multicolumn{2}{l}{\textbf{Decoding Time and Space}} \\
\toprule
Attention & $\sigma(\mQ\mK^\top)$ 
& Mask (\ding{55}) & FlashAttention $\mathcal{O}(T^2)$ & $\mathcal{O}(T)$ & $\mathcal{O}(T)$ \\
Linear Attention & $\mQ\mK^\top$ & Mask (\ding{55}) & Chunk-recurrent $\mathcal{O}(T)$ &  $\mathcal{O}(1)$ & $\mathcal{O}(1)$ \\
RetNet & $\mQ\mK^\top$ & Semiseparable (\ding{55}) & Chunk-recurrent $\mathcal{O}(T)$ & $\mathcal{O}(1)$ & $\mathcal{O}(1)$ \\
Mamba-2 & $\mQ\mK^\top$ & Semiseparable (\ding{51}) & Chunk-recurrent $\mathcal{O}(T)$ & $\mathcal{O}(1)$ & $\mathcal{O}(1)$\\
Multi-Hyena & $\mQ\mK^\top$ & Toeplitz (\ding{55}) & FFT $\mathcal{O}(T \log T)$ &  $\mathcal{O}(\log^2 T)$ & $\mathcal{O}(T)$ \\
DeltaNet & $\mathcal{T}_{\mK}(\mQ\mK^\top)$ & Mask (\ding{55}) & Chunk-recurrent $\mathcal{O}(T)$ & $\mathcal{O}(1)$ & $\mathcal{O}(1)$ \\
Gated DeltaNet & $\mathcal{T}_{\mK}(\mQ\mK^\top)$ & Semiseparable (\ding{51}) & Chunk-recurrent $\mathcal{O}(T)$ & $\mathcal{O}(1)$ & $\mathcal{O}(1)$\\
\midrule
Log-Linear Mamba-2 & $\mQ\mK^\top$ &  Hierarchical (\ding{51}) & Chunk-scan $\mathcal{O}(T \log T)$ &  $\mathcal{O}(\log T)$ & $\mathcal{O}(\log T)$  \\
Log-Linear Gated DeltaNet  & $\mathcal{T}_{\mK}(\mQ\mK^\top)$ &  Hierarchical (\ding{51}) & Chunk-scan $\mathcal{O}(T \log T)$ &  $\mathcal{O}(\log T)$ & $\mathcal{O}(\log T)$ \\
\bottomrule
\end{tabular}
\vspace{-1mm}
\caption{Summary of efficient attention mechanisms under the unified formulation: $\mP=\mA\odot \mM, \mO = \mP\mV$. $\mM$ is a lower-triangle (causal) matrix.  We use symbol $\mathcal{T}_{\mK}\left({\mA}\right) = \left(\mA \odot \mL \right)\left(\mI + \mK\mK^\top \odot (\mI - \mL) \right)^{-1}$ for notational brevity, where $\mL$ is a lower-triangular matrix of 1s. Here decoding time is the time per step, and decoding space refers to the overall memory complexity during generation.}
\vspace{-5mm}
\label{tab:efficient-attn}
\end{table}

Given an input sequence of length $T$ and the corresponding key, query, value matrices $\mK, \mQ, \mV~\in~\sR^{T \times d}$,  softmax attention obtains the output $\mO \in \sR^{T \times d}$ for all time steps via  $\mO = \operatorname{softmax}(\mQ \mK^\top \odot \mM) \mV$, where $\mM \in \{-\infty, 0\}^{T \times T}$ is a causal masking matrix.  This incurs $\gO(T^2)$ compute and $\gO(T)$ memory, which makes it costly to apply to long sequences. As a response, there has been much recent work on  efficient  alternatives with sub-quadratic compute and sub-linear memory, including linear attention, state-space models, and long convolution models. Despite their  differences, many of these approaches can be captured by the following equation:
\begin{equation}
\mP = \mA \odot \mM, \quad \mO = \mP \mV,
\label{eq:efficient-attention-as-structured-matrices}
\end{equation}
where $\mA \in \mathbb{R}^{T \times T}$ is an attention-like matrix (e.g., $\mQ\mK^\top$ in the case of ordinary linear attention) and $\mM\in \mathbb{R}^{T \times T}$ is a lower-triangular causal masking matrix (e.g., $\mM \in \{0,1\}^{T\times T}$ for linear attention).
By separating out the interaction terms $\mA$ and the (potentially data-dependent) masking matrix $\mM$, this abstraction reveals commonalities across a broad class of models, as shown in Table~\ref{tab:efficient-attn}. Different structures imposed on $\mM$  can lead to  efficient training and inference algorithms. We now describe key models that fit within this framework.

\paragraph{Linear attention.} Linear attention \cite{katharopoulos2020transformers} simply removes  the softmax operation, resulting in the following parallel form\footnote{Here we work linear attention without any feature maps or normalization, since most recent works have found them to be unnecessary (although see \citep{kacham2023polysketchformer,buckman2024,arora2024simple}).}
\begin{align*}
  \mO = (\mQ\mK^\top \odot \mM) \, \mV, \quad \mM_{ij} = \mathbf{1}\{i \le j\}.
\end{align*}
Linear attention can be reparameterized into the following ``recurrent form'' for inference,
\begin{align*}
\mS_t = \mS_{t-1} + \vv_t \vk_t^\top, \quad  \vo_t = \mS_t \vq_t,
\end{align*}
which enables linear-time constant-memory sequence modeling. 

\paragraph{Linear attention with (data-dependent) gates.} 
Vanilla linear attention lacks a forgetting mechanism, which has been shown to be crucial in ordinary RNNs. One way to incorporate such a mechanism is through a scalar gate $\alpha_t \in (0,1)$, which results in recurrence $\mS_t = \alpha_t \mS_{t-1} + \vv_t \vk_t^\top$.
This has the following corresponding parallel form:
\begin{align}
  \mO = (\mQ\mK^\top \odot \mM) \mV, \quad \mM_{ij} = \prod_{k=j+1}^i \alpha_k.
\label{eq:m-sss}
\end{align}
Originally introduced by \citet{peng_random_2021},  gated linear attention  has enjoyed a resurgence in recent years \citep{qin_hgrn2_2024,peng2024eagle,yang2023gated,katsch2023gateloop} and are an instance of time-varying SSMs \citep{Gu2023MambaLS,dao2024transformers}. Well-known models in this family include RetNet \citep{sun2023retentive}, which uses a data-\emph{in}dependent gate $\alpha_t = \alpha$, and Mamba-2 \citep{dao2024transformers}, which uses the above data-dependent gate. \citet{dao2024transformers} show that with a scalar gating factor, $\mM$ has a 1-semiseparable structure where every submatrix in the lower triangular portion has rank at most 1, which can enable efficient training.

\paragraph{Linear attention with the delta rule.} DeltaNet \citep{schlag_linear_2021}  is a type of linear attention layer which updates the hidden state via the delta rule \citep{widrow_adaptive_1988},\footnote{Linear attention with the delta rule is also an instance of a fast-weight programmer \citep{schmidhuber1992learning}.} where the recurrent form is given by\footnote{The actual DeltaNet recurrence is given by $\mS_t = \mS_{t-1}(\mI - \beta_t \vk_t \vk_t^\top) + \vv_t \vk_t^\top$ where $\beta_t$ is a data-dependent scalar value   in either $(0,1)$ or $(0,2)$, but we set $\beta_t=1$ here for notational brevity.} 
\begin{align*}
\mS_t = \mS_{t-1}\left(\mI - \vk_t \vk_t^\top\right) + \vv_t \vk_t^\top, \quad  \vo_t = \mS_t \vq_t.
\end{align*}
While the original work used a purely recurrent form, \citet{yang2024parallelizing} recently show that it is possible to parallelize DeltaNet across sequence length through leveraging a compact representation of Householder matrices \citep{bischof_wy_1985,Joffrain2006AccumulatingHT}, resulting in the 
following parallel form (cf. \cite[\S 3.2]{yang2024parallelizing}):
\begin{align*}
\mO = \left(\underbrace{\left(\mQ\mK^\top \odot \mL\right)\left(\mI + \mK\mK^\top \odot (\mL - \mI )\right)^{-1}}_{\mA} \odot \mM\right) \mV
\end{align*}
where $\mL$ and $\mM$ are lower-triangular matrices consisting of 1s. Since $\mA$ itself is already lower-triangular, the causal masking matrix $\mM$ is not strictly necessary in the above. However, by changing $\mM$ to have 1-semiseparable structure as in Mamba-2, we can recover Gated DeltaNet \citep{yang2024gated}, whose recurrence is given by $\rmS_t = \alpha_t \rmS_{t-1}(\mI - \vk_t \vk_t^\top) + \vv_t \vk_t^\top$. Linear attention with such data-dependent  structured transition matrices has been shown to be theoretically more expressive than linear attention with multiplicative gates when it comes to certain types of \emph{state-tracking} tasks \citep{merrill2024illusion,grazzi2025unlocking,siems2025deltaproduct,peng2025rwkv}, which make these layers attractive targets to generalize via our log-linear attention framework.

\paragraph{Long convolution models.}
 Long-convolution sequence models, where the convolution kernel size equals the sequence length, can also be cast into this framework. For example, Toeplitz neural network \citep{qin2023toeplitz} and MultiHyena \cite{massaroli2023laughing} layers are given by 
$\mO = (\mQ\mK^\top \odot \mT_h) \, \mV$, where  $\mT_h$ is a causal Toeplitz matrix generated by a long convolution kernel $\vh \in \mathbb{R}^T$, i.e.,  $\mT_h[i,j] = \vh[i-j]$ for $i \geq j$ and 0 otherwise. Other long convolutional variants like H3 \citep{fu2023hungry} and Hyena \citep{hyena} also admit a precise attention-style formulation \citep{massaroli2023laughing}.
While the decoding speed of long convolution models can be  improved from $\mathcal{O}(T)$ to $\mathcal{O}(\log^2 T)$ per step \citep{oncescu2025flash},  their memory cost remains linear, i.e., the same as in softmax attention.  However, some long convolution models such as S4 \citep{gu2022efficiently} admit a reparameterization into a time-invariant SSM and thus enjoy constant-memory inference. There has also been efforts to distill long convolution models into RNNs \citep{massaroli2023laughing,qin-zhong-2023-accelerating}, but these inherit the memory bottleneck of RNNs.

 \paragraph{Relationship between masking structure and efficient algorithms.} Using an unstructured $\mM$ (e.g., a random lower-triangular matrix) degrades both compute and memory complexity to softmax attention-levels, despite the absence of softmax; i.e.,  the \emph{structure} of $\mM$ is essential for training/inference efficiency, not just the removal of  softmax. In linear attention where $\mM$ is a lower-triangular matrix of 1's, we can compute $\mO$ chunkwise, leading to an $\mathcal{O}(T)$ algorithm.\footnote{This  algorithm  depends on the chunk size $C$, but since $C$ is a hyperparameter this is still linear in $T$.} This algorithm generalizes to the gated case where $\mM$ has 1-semiseparable structure as shown in \citep{dao2024transformers}. Long convolution models can 
 use FFT to bring down the cost to $\mathcal{O}(T \log T)$.

\section{Log-Linear Attention}
\label{sec:log-linear-attention}
\begin{wrapfigure}{r}{0.43\linewidth}
    \vspace{-12mm}
    \centering\includegraphics[width=0.99\linewidth]{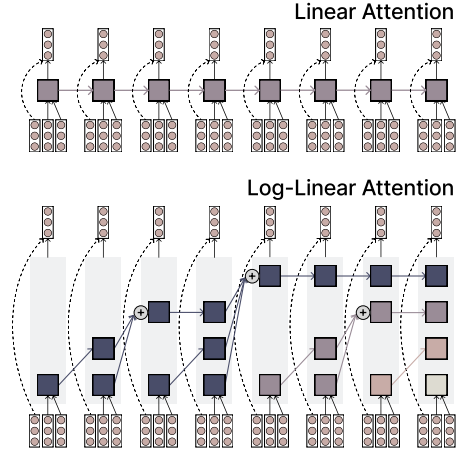} %
    \caption{Standard linear attention (top) vs. log-linear attention (bottom). The input consists of query, key, and value vectors.}
    \label{fig:recurrent}
\vspace{-5mm}
\end{wrapfigure}
The previous section showed that the structure of the masking matrix $\mM$ determines how compute and memory scale with sequence length. Semiseparable structures cover many efficient architectures, yielding $\mathcal{O}(T)$ training time and $\mathcal{O}(1)$ decoding memory. This motivates two questions: \emph{(i)} what additional structures allow greater flexibility while retaining subquadratic training complexity, and \emph{(ii)} can such models admit a recurrent form with \emph{sublinear} decoding memory?

We answer both by introducing \emph{log-linear attention}, which shapes $\mM$ to achieve $\mathcal{O}(T\log T)$ computation and $\mathcal{O}(\log T)$ memory. Concretely, log-linear attention replaces the semiseparable mask with a \emph{hierarchical} one, extending linear attention beyond semiseparable temporal structure and accommodating a broader class  structures for $\mA$. As case studies, we instantiate log-linear variants of Mamba-2 and Gated DeltaNet.

During decoding, log-linear attention employs a Fenwick tree scheme~\citep{Fenwick1994AND} that partitions inputs into power-of-two segments. Each position summarizes its prefix, enabling queries to attend to $\mathcal{O}(\log T)$ hidden states across multiple scales (Fig.~\ref{fig:recurrent}). This design preserves fine-grained access to recent tokens while requiring only $\mathcal{O}(\log T)$ time and memory. 
We first focus on the simplest form of linear attention (without gating) in   \S~\ref{subsec:fenwick-tree} and show how log-linear attention extends it by maintaining independent recurrent states across temporal segments. Practical gated variants are presented in \S~\ref{sect:loglinear-variant}.

\subsection{Fenwick Tree Partitioning and Hierarchical Matrices}
\label{subsec:fenwick-tree}

\begin{wrapfigure}{r}{0.4\linewidth}
\vspace{-4mm}
\centering
\includegraphics[width=0.95\linewidth]{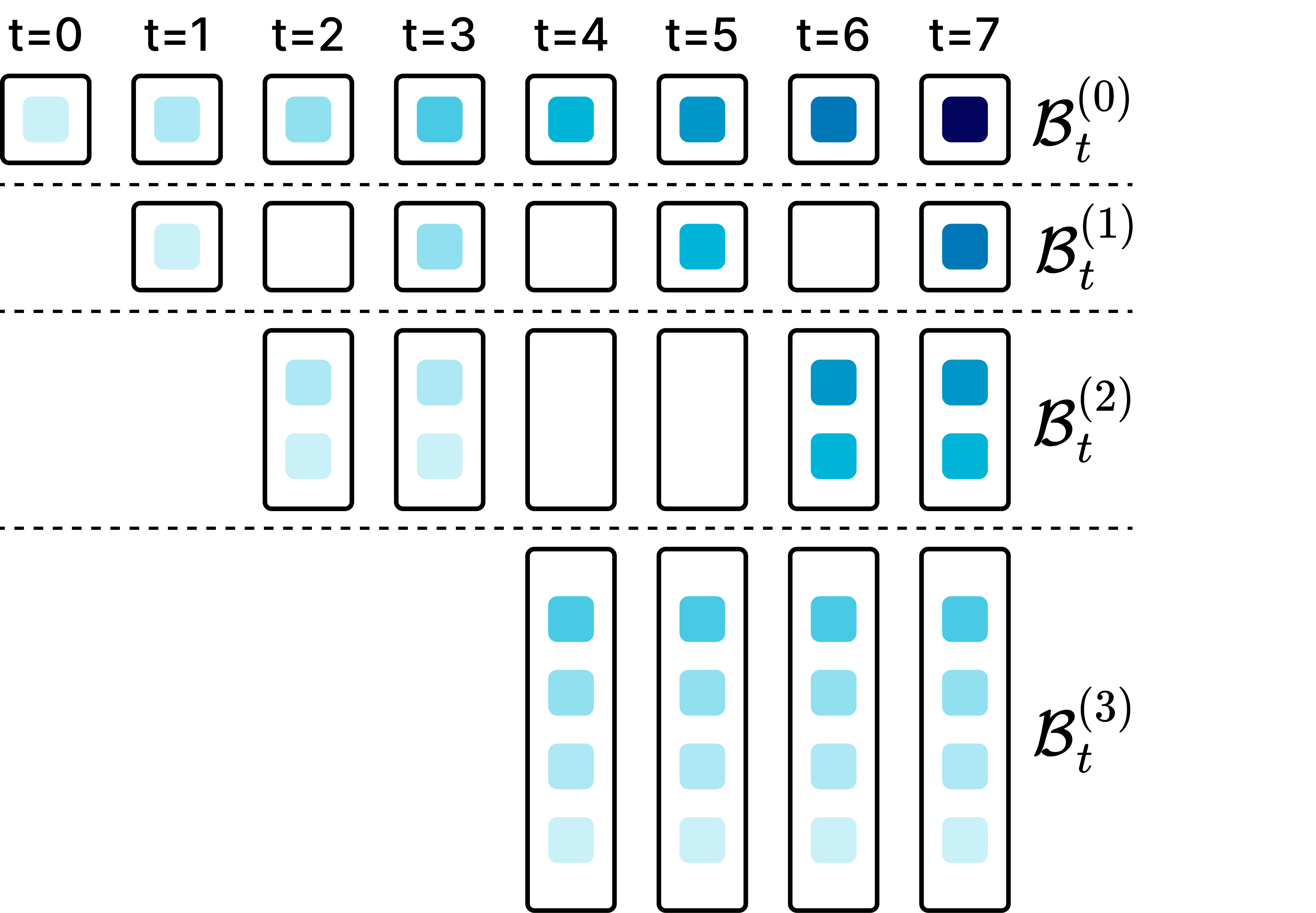}
\vspace{-1mm}
\captionof{figure}{Fenwick tree bucket assignments.}
\label{fig:bucket-assignments}
\vspace{-5mm}
\end{wrapfigure}

From a decoding perspective, attention can be viewed as a mechanism that partitions the prefix $[0, t)$ into a set of buckets, each summarizing a portion of the past. In vanilla attention, every token forms its own bucket, resulting in $t$ buckets of size $1$, each stored as a fixed-size state (the KV caches. At the other extreme, linear attention (and state-space models) aggregates the entire prefix into a single bucket of size $t$, again represented by a fixed-size state.

Log-linear attention strikes a balance by partitioning the prefix into buckets of exponentially increasing size via a Fenwick-tree decomposition~\citep{Ryabko1992AFO,Fenwick1994AND}. This induces a natural inductive bias: recent tokens are retained at high resolution, while more distant tokens are summarized more coarsely. The partition contains at most
$L=\mathcal{O}(\log T)$ disjoint buckets indexed by level $\ell$.%
\footnote{More precisely, this divides the prefix $[0, t)$ into up to $L = \lceil \log_2 t + 1 \rceil + 1$ disjoint buckets. This decomposition is guided by the function $\operatorname{lssb}(t) = \max{\{\ell \in \mathbb{N} \mid 2^\ell \text{ divides } t\}}$, which identifies the least significant set bit in the binary representation of $t$. Conceptually, the partitioning proceeds greedily, at each step subtracting the largest power of two that fits within the remaining segment of the prefix,\vspace{-2mm}
\begin{equation*}
\begin{aligned}
&b_t^{(i)} {=} 
\begin{cases} 
t & \text{if } i = 0 \\[1pt]
b^{(i-1)}_t {-}2^{\operatorname{lssb}\left(b^{(i-1)}_t\right)} & \text{otherwise}
\end{cases},
&\mathcal{B}_t^{(\ell)} {=}
\begin{cases} 
\{b_t^{(0)}\} & \text{if } \ell = 0 \\[1pt]
\{b^{(i + 1)}_t, \cdots, b^{(i)}_t {-} 1\} & \text{if } \ell = \operatorname{lssb}\left(b^{(i)}_t\right) {+} 1 \\[1pt]
 \varnothing & \text{otherwise}
\end{cases}
\end{aligned} 
\end{equation*}
}
Each bucket $\mathcal{B}_t^{(\ell)}$ has size $|\mathcal{B}_t^{(\ell)}| = 2^{\ell - 1}$ for $\ell \ge 1$, plus a sentinel bucket $\mathcal{B}_t^{(0)}$ of size $1$. See Fig.~\ref{fig:bucket-assignments} for an illustration.

Log-linear attention maintains a separate recurrent memory $\mS_t^{(\ell)} \in \mathbb{R}^{d \times d}$ for each bucket. At time $t$, the contribution of bucket $\ell$ to the output is weighted by a nonnegative coefficient $\lambda_t^{(\ell)}$, parameterized as a linear function of the current input $\vx_t$. This allows the model to adaptively emphasize different temporal scales. The output is computed as,
\begin{equation}
\vo_t 
= \sum_{\ell = 0}^{L-1} \lambda_t^{(\ell)} \, \vq_t^\top 
\Bigg(\sum_{s \in \mathcal{B}_t^{(\ell)}} \vv_s \vk_s^\top\Bigg) 
= \sum_{\ell = 0}^{L-1} \lambda_t^{(\ell)} \, \vq_t^\top \mS_t^{(\ell)}.
\label{eq:fenwick-tree-partitioning}
\end{equation}
We observe that when all $\lambda_t^{(\ell)}$ are the same (or more generally when the $\lambda_t^{(\ell)}$ and $\lambda_t^{(\ell^\prime)}$ are linearly related across time) log-linear attention collapses to linear attention. Allowing distinct $\lambda_t^{(\ell)}$ is therefore essential for capturing multi-scale temporal structure.

\paragraph{Parallel form.}
The recurrent form in Eq.~\ref{eq:fenwick-tree-partitioning} is conceptually simple but inefficient on modern accelerators, which are optimized for high-throughput matrix–matrix multiplication. To leverage this hardware and enable parallelization across time, we reformulate the expression in a matrix-multiplication–friendly form as in \S\ref{sec:efficient_attention_with_structured_matrices}:
\begin{align}
\mO=\underbrace{\left(\mQ \mK^{\top} \odot \mM^{\mathcal{H}}\right)}_{\mP} \mV, \quad
\mM^{\mathcal{H}}_{ts} =
\begin{cases}
\lambda_t^{\ell(t,s)} & \text{if } s \leq t, \\
0 & \text{otherwise},
\end{cases}
\label{eq:h-parallel-form}
\end{align}
where $\ell(t,s)$ denotes the bucket level of token $s$ relative to time $t$ under Fenwick-tree partitioning. For readability, we omit explicit $(t,s)$ indices when unambiguous.
The matrix $\mP$ is a hierarchical matrix which inherits structured low-rank pattern from the hierarchical partitioning, given below. In \S\ref{sec:parallel-algorithm}, we exploit this structure to design a parallel training algorithm with $\mathcal{O}(T \log T)$ complexity.

\begin{equation*}
\renewcommand{\arraystretch}{1.5}
\scriptsize
\begin{bNiceArray}{cc|cc|cc|cc}
\lambdalevel{0}{0} \vq_0^\top \vk_0 & & & & & & & \\
\lambdalevel{1}{1} \vq_1^\top \vk_0 & \lambdalevel{0}{1} \vq_1^\top \vk_1 & & & & & & \\
\midrule
\Block[fill=level2!15]{2-2}{
\begin{bmatrix}
\lambdalevel{2}{2} \vq_2 \\
\lambdalevel{2}{3} \vq_3
\end{bmatrix}
\begin{bmatrix}
\vk_0 \\
\vk_1
\end{bmatrix}^{\top}
} & & \lambdalevel{0}{2} \vq_2^\top \vk_2 & & & & & \\
&& \lambdalevel{1}{3} \vq_3^\top \vk_2 & \lambdalevel{0}{3} \vq_3^\top \vk_3 & & & & \\
\midrule
\Block[fill=level3!15]{4-4}{
\begin{bmatrix}
\lambdalevel{3}{4} \vq_4 \\
\lambdalevel{3}{5} \vq_5 \\
\lambdalevel{3}{6} \vq_6 \\
\lambdalevel{3}{7} \vq_7
\end{bmatrix}
\begin{bmatrix}
\vk_0 \\
\vk_2 \\
\vk_3 \\
\vk_1
\end{bmatrix}^{\top}
} & & & & \lambdalevel{0}{4} \vq_4^\top \vk_4 & & & \\
&&&& \lambdalevel{1}{5} \vq_5^\top \vk_4 & \lambdalevel{0}{5} \vq_5^\top \vk_5 & & \\
\cmidrule{5-8}
&&&& 
\Block[fill=level2!15]{2-2}{
\begin{bmatrix}
\lambdalevel{2}{6} \vq_6 \\
\lambdalevel{2}{7} \vq_7
\end{bmatrix}
\begin{bmatrix}
\vk_4 \\
\vk_5
\end{bmatrix}^{\top}
} & & \lambdalevel{0}{6} \vq_6^\top \vk_6 & \\
&&&&&& \lambdalevel{1}{7} \vq_7^\top \vk_6 & \lambdalevel{0}{7} \vq_7^\top \vk_7 \\
\end{bNiceArray}
\end{equation*}

\paragraph{Remark.}
The matrix $\mM^{\mathcal{H}}$ (and $\mA$) is a lower-triangular instance of a hierarchical ($\mathcal{H}$) matrix—specifically, of the HODLR (Hierarchically Off-Diagonal Low-Rank) type. When constructed using schemes like the Fenwick tree, it inherits the recursive partitioning and low-rank off-diagonal blocks that define $\mathcal{H}$ matrices.
This establishes a direct connection between log-linear attention and hierarchical matrices: the attention operator corresponds to structured matrix multiplication with an $\mathcal{H}$ matrix. We refer to $\mM^{\mathcal{H}}$ as a quasi-$\mathcal{H}$ matrix—a specialized class lying between general $\mathcal{H}$ and semiseparable matrices, designed to support $\mathcal{O}(\log T)$-space recurrence. See Section~\ref{appendix:quasi-hierarchical} for details.

\vspace{-2mm}
\subsection{Memory-efficient decoding}
\vspace{-2mm}

Let $\operatorname{lssb}(t)$ denote the index of the least significant set bit in the binary representation of $t$.  
The states $\{\mS_t^{(\ell)}\}_{\ell}$ evolve according to the following recurrence (using linear attention for simplicity):

\begin{wrapfigure}{l}{0.3\textwidth}
\vspace{-3mm}
\begin{align*}
\small
\mS_{t}^{(\ell)} {=} 
\begin{cases} 
\vv_t \vk_t^\top & \text{if } \ell {=} 0 \\[1pt]
0 & \text{if } 0 {<} \ell {\leq} \operatorname{lssb}(t) \\[1pt]
\sum_{\ell^\prime = 0}^{\ell - 1} \mS_{t-1}^{(\ell^\prime)} & \text{if } \ell {=} \operatorname{lssb}(t) {+} 1 \\[1pt]
\mS_{t-1}^{(\ell)} & \text{if } \ell {>} \operatorname{lssb}(t) {+} 1
\end{cases}
\end{align*}
\vspace{-3mm}
\end{wrapfigure}
At each step, the immediate term $\vv_t \vk_t^\top$ enters the finest level; buckets up to $\operatorname{lssb}(t)$ merge and promote one level coarser. When $t$ is a power of two the hierarchy expands by one bucket. This Fenwick-like organization enables online processing with $\mathcal{O}(\log T)$ memory while retaining efficient multiscale access.

\subsection{Efficient Algorithm for Training}
\label{sec:parallel-algorithm}
Chunkwise parallelism for linear attention~\citep{sun2023retentive,yang2023gated,dao2024transformers} partitions a sequence of length $T$ into chunks of size $C$, which are processed in parallel while exchanging only limited information across boundaries. This approach balances two extremes: it avoids the prohibitive cost of global attention while exposing substantially more parallelism than purely recurrent execution. We extend this idea to the log-linear setting and develop an efficient \emph{chunkwise} training algorithm.

\begin{figure}[t!]
\vspace{-6mm}
\centering
\includegraphics[width=0.87\textwidth]{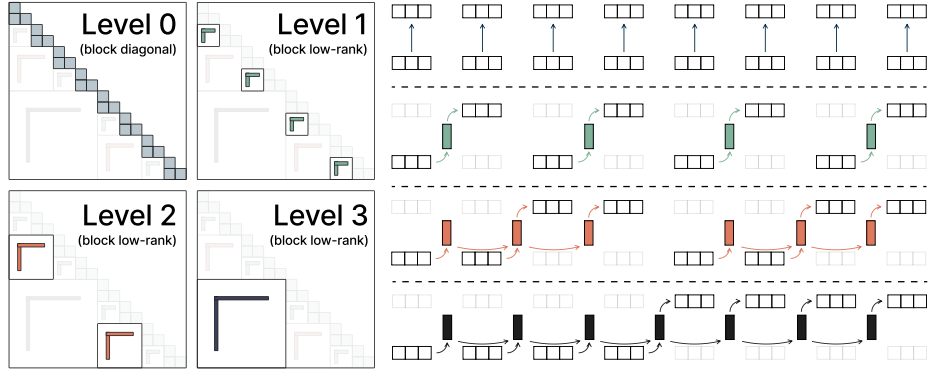}
\caption{
\textbf{Left}: Decomposition of the matrix $\mM^{\mathcal{H}}$.  
\textbf{Right}: Chunkwise algorithm (Algorithm~\ref{alg:hattention}). Level 0 handles intra-chunk computations using a quadratic (in chunk size) algorithm, which is efficient due to small chunk sizes. Levels 1 and above perform inter-chunk computations by invoking existing inter-chunk primitives multiple times, with overall complexity logarithmic in the number of chunks.
}
\label{fig:h-matrices}
\end{figure}

For a given chunk size $C$, the matrix $\mM^{\mathcal{H}}$ admits the structured decomposition,
\begin{equation}
\mM^{\mathcal{H}} = \mD + \sum_{\ell=\ell_C}^{L-1} \mM^{(\ell)}, \quad
\mM^{(\ell)}_{ts} =
\begin{dcases*}
\lambda_{t}^{(\ell)} \mM^{\mathcal{S}}_{ts}, & if $ s \in \mathcal{B}_t^{(\ell)} $,\\
0, & otherwise. 
\end{dcases*}
\label{eq:h-decomposition}
\end{equation}
where $\mD$ is block-diagonal with $\tfrac{T}{C}$ causal blocks $\{\mD^{[k]}\}$ of size $(C {\times} C)$, capturing intra-chunk interactions via $(\mD^{[i]})_{ts}=\lambda_{iC+t}^{(\ell)}\,\mM^{\mathcal{S}}_{ts}$. The remaining $\{\mM^{(\ell)}\}$ encode inter-chunk dependencies in blockwise low-rank form. Indexing begins at $\ell_C$, the level aligned to chunk size $C$; levels $\ell<\ell_C$ collapse into $\mD$ (Fig.~\ref{fig:h-matrices}, left).

Building on this structure, we propose a chunkwise algorithm for log-linear attention (Algorithm~\ref{alg:hattention}). As summarized in Fig.~\ref{fig:h-matrices} (right), the method introduces only a logarithmic overhead compared with standard linear attention. Computation proceeds in two stages:

\textbf{Intra-chunk stage ($\ell < \ell_C$).}  
The block-diagonal component $\mD$ is treated as a dense matrix within each chunk. Each block costs $\mathcal{O}(C^2)$, giving a total complexity of $\mathcal{O}(TC)$.

\textbf{Inter-chunk stage ($\ell \geq \ell_C$).}  
The matrices $\{\mM^{(\ell)}\}$ reduce to scaled sequentially semi-separable structures (Eq.~\ref{eq:h-decomposition}). With efficient state-passing primitives (e.g., Mamba-2, Gated DeltaNet), inter-chunk dependencies are computed using only $\mathcal{O}\!\left(\log \tfrac{T}{C}\right)$ primitive calls. Each call requires $\mathcal{O}(T)$ time and memory,\footnote{At level $\ell$, $\mM^{(\ell)}$ contains $\tfrac{T}{2^{\ell-1} C}$ chunks of size $2^{\ell-1} C$. Redundant work can be avoided, reducing cost by a constant factor of two.} leading to an overall complexity of $\mathcal{O}(T \log \tfrac{T}{C})$.

Our algorithm extends the classical parallel prefix-sum (scan) to a hierarchical setting—a \emph{chunkwise parallel scan}. Unlike token-level scans, which often suffer from memory-bandwidth bottlenecks during training~\citep{yang2023gated}, the chunkwise formulation reorganizes recurrent updates into parallel chunk operations. Concretely, it executes $\mathcal{O}(\log T)$ independent scans (one per memory level), each implementable with standard methods such as the Blelloch scan~\citep{blelloch1990prefix}. Layer-specific weights (e.g., $\lambda_t^{(\ell)}$) can easily be incorporated into these scans.

\subsection{Log-Linear Variants of Mamba-2 and Gated DeltaNet}
\label{sect:loglinear-variant}

We next apply the above construction to Mamba-2~\cite{dao2024transformers} and Gated DeltaNet~\cite{yang2024gated}. As discussed in \S\ref{sec:efficient_attention_with_structured_matrices}, both models use gating mechanisms that induce a sequentially semiseparable (SSS) temporal structure in the mask $\mM^{\mathcal{S}}$ (with $\mM_{ij}=\prod_{k=j+1}^{i}\alpha_k$; see Eq.~\ref{eq:m-sss}). The two architectures differ in how they parameterize the transition matrix $\mA$.

Our approach preserves the original form of $\mA$ in each model while composing the attention mask with its log-linear variant $\mM = \mM^{\mathcal{S}} \odot \mM^{\mathcal{H}}$.\footnote{More precisely, the elementwise product of an SSS matrix and an $\mathcal{H}$ matrix remains an $\mathcal{H}$ matrix. We separate them here for clarity.} We refer to the resulting models as \emph{log-linear} Mamba-2 and \emph{log-linear} Gated DeltaNet. Their parallel forms are given by,
\begin{align*}
    \mO &= \left(\mQ\mK^T \odot \mM^{\mathcal{S}} \odot \mM^{\mathcal{H}}\right) \, \mV && \textit{Log-Linear Mamba-2} \\
    \mO &= \left(\left(\mQ\mK^\top \odot \mL \right) \left(\mI+\mK\mK^\top \odot \left(\mL - \mI \right)\right)^{-1} \odot \mM^{\mathcal{S}} \odot \mM^{\mathcal{H}}\right)\,\mV && \textit{{Log-Linear} Gated DeltaNet} 
\end{align*}
More broadly, any linear-attention mechanism with structured memory and an efficient chunkwise-parallel primitive can be “lifted’’ to a log-linear variant by composing its temporal mask with $\mM^{\mathcal{H}}$.

\subsection{Implementation}
\begin{figure}[t] %
\centering
\vspace{-15pt}
\includegraphics[width=0.85\linewidth]{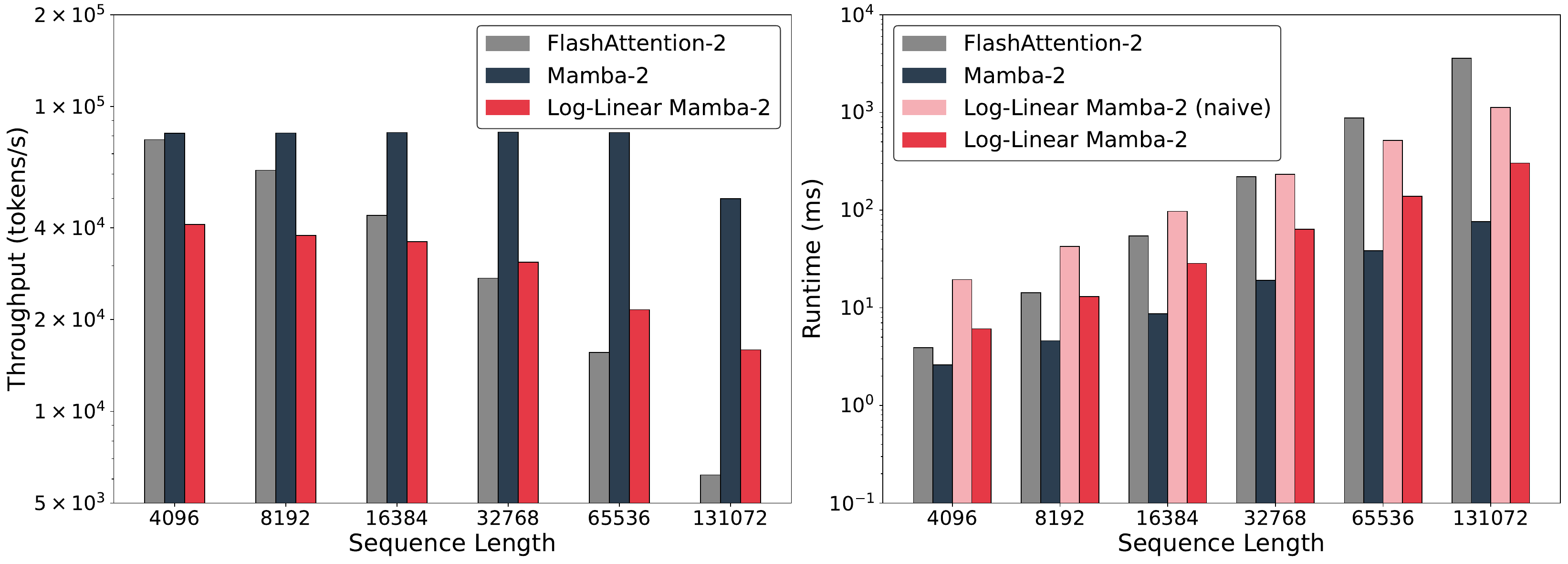}
\vspace{-7pt}
\captionof{figure}{
Training throughput (left; higher is better) and kernel runtime for a forward and backward pass (right; lower is better) across varying sequence lengths. \textbf{Log-Linear Mamba-2 (naive)} denotes repeated application of the existing Mamba-2 primitives, while \textbf{Log-Linear Mamba-2} uses a custom implementation with optimizations such as level fusion. The throughput drop at sequence length 131K is due to gradient checkpointing to reduce memory usage.
Experiments were run on an H100 GPU with batch size 2, 48 heads, head dimension 64, state dimension 128, and chunk size 64. We use MVA for (Log-Linear) Mamba-2, and GQA for FlashAttention-2.}
\vspace{-5mm}
\label{fig:impl}
\end{figure}

We implemented the chunkwise parallel scan algorithm in \texttt{Triton}~\citep{tillet2019triton}. The custom kernel for {log-linear} Mamba-2 outperforms FlashAttention-2~\citep{dao2023flashattention2} (forward + backward) at sequence lengths beyond 8K. In full training setups, throughput depends on model architecture. Notably, {log-linear} Mamba-2 (with MLP) surpasses Transformer throughput at 32K, despite additional layers like depthwise convolutions absent in the Transformer. See Fig.~\ref{fig:impl} and Sec.~\ref{appendix-subsec:implementations} for details.

\section{Experiments}
We conduct a suite of experiments across both synthetic and real-world benchmarks. We emphasize that our experiments are not necessarily intended position log-linear attention as the best subquadratic architecture, but rather to highlight the promise of our framework compared to sensible baselines.
\subsection{Synthetic Benchmark}

\begin{wraptable}{r}{0.529\textwidth}
\centering
\small
\begin{tabular}{r|rrr}
\toprule
\textbf{Dimension} & \textbf{16} & \textbf{32} & \textbf{64} \\
\midrule
Transformer          & ${\ge}$ 99 & ${\ge}$ 99 & ${\ge}$ 99 \\
Mamba-2              & 46.9 (2.3) &  75.1 (4.9) &  89.6 (6.1) \\
\rowcolor{black!10}
w/ \emph{Log-Linear} & 55.9 (9.1) & 76.5 (4.8) & 92.9 (2.7) \\
Gated DeltaNet       & 38.4 (1.0) & 79.0 (2.1) & ${\ge}$ 99 \\
\rowcolor{black!10}
w/ \emph{Log-Linear} & 40.0 (1.4) & 84.4 (1.2) & ${\ge}$ 99 \\
\bottomrule
\end{tabular}
\centering
\vspace{-2mm}
\caption{Average accuracies and standard deviations (in parentheses) on MQAR over 5 seeds. Training was early stopped when accuracy exceeded 99\%.}
\label{tab:mqar}
\vspace{-3.9mm}
\end{wraptable}

We begin by evaluating models on the multi-query associative recall (MQAR) task~\citep{arora2023zoology}, a standard diagnostic benchmark for testing in-context recall. Our setup closely follows~\citet{arora2024simple}: models are trained and evaluated on 256-token sequences containing 4 to 64 key-value pairs (excluding the length generalization component), with tuned learning rates. For log-linear models, we also tune the $\lambda$ parameterization. We run each configuration with five seeds. Training was early stopped when accuracy exceeded 99\%. Additional experimental details are provided in \S\ref{appendix-sec:experiment-details}. 
As shown in Table~\ref{tab:mqar}, log-linear attention performs well—even when applied on top of associative recall-optimized models like Gated DeltaNet.

\subsection{Language Modeling}

We perform academic-scale language modeling pretraining from scratch using 50B tokens on the Long-Data-Collections  dataset,\footnote{\url{https://huggingface.co/datasets/togethercomputer/Long-Data-Collections}.} using a sequence length of 16K. All models have 21 layers and use a hidden size of 1536. We use a Transformer with 16 attention heads and a RoPE base of 500K, a modified Mamba-2 with 48 heads and MLP layers, and a Gated DeltaNet with 6 heads.  The Transformer, Mamba-2, and Gated DeltaNet models contain 693M, 802M, and 793M parameters, respectively. For the \emph{log-linear} variants, we apply a linear layer on top of the hidden states to compute the per-head values $\lambda^{(\ell)}_t$. This adds less than $3\%$ additional parameters for Mamba-2 (825M) and less than $0.4\%$ for Gated DeltaNet (796M). Since Mamba-2 and Gated DeltaNet have more parameters than ordinary Transformers, we also include a (roughly) parameter-matched Transformer variant with 24 layers (778M parameters) for comparison.
For our log-linear variants, we use the default hyperparameters from the baselines (\S\ref{appendix-sec:experiment-details}).
We also evaluated a parameter-matched Hyena model~\cite{hyena}, which also has log-linear compute (but linear memory). As its WikiText perplexity (around $29$) was substantially higher than that of the other models (${<}23$), our main experiments focus on the Transformer, Mamba-2, and Gated DeltaNet families.

\begin{wraptable}{r}{0.5\textwidth}
\vspace{-3.5mm}
\centering
\small
\begin{tabular}{l|ccc}
\toprule
\textbf{Model}  & \textbf{Wiki.}  &  \textbf{LMB.} &  \textbf{LMEval}  \\
 & ppl $\downarrow$  &  ppl $\downarrow$  & average $\uparrow$ \\
\midrule
Transformer          & 21.56 & 22.14 & 44.0 \\
w/ \emph{24 Layers}  & 21.13 & 21.17 & 45.6 \\
Hyena                & 29.50 & /     & /    \\
Mamba-2              & 22.44 & 24.14 & 44.8 \\
\rowcolor{black!10}
w/ \emph{Log-Linear} & 22.11 & 21.86 & 44.9 \\
Gated DeltaNet       & 21.73 & 19.71 & 45.0 \\
\rowcolor{black!10}
w/ \emph{Log-Linear} & 21.45 & 18.09 & 45.5 \\
\bottomrule
\end{tabular}
\centering
\vspace{-2mm}
\caption{PPL and commonsense reasoning.} 
\vspace{-5mm}
\label{tab:lmeval-small}
\end{wraptable}

\paragraph{Standard benchmarks.}
Following prior work~\citep{dao2024transformers,yang2024gated}, we evaluate models on WikiText perplexity and several zero-shot commonsense reasoning benchmarks (Table~\ref{tab:lmeval}). These are short-context tasks and are therefore largely insensitive to model state size. As such, we generally expect the log-linear variants to perform comparably to their linear counterparts. Log-Linear Mamba-2 improves upon its linear counterpart in perplexity and in half of the commonsense reasoning tasks. Log-Linear Gated DeltaNet shows stronger gains, outperforming its linear version in perplexity and in all but one reasoning benchmark. Notably, it also outperforms a layer-matched Transformer across all metrics and a parameter-matched Transformer on half of them.
\begin{figure}
\vspace{-15pt}
\centering
\includegraphics[width=0.99\linewidth]{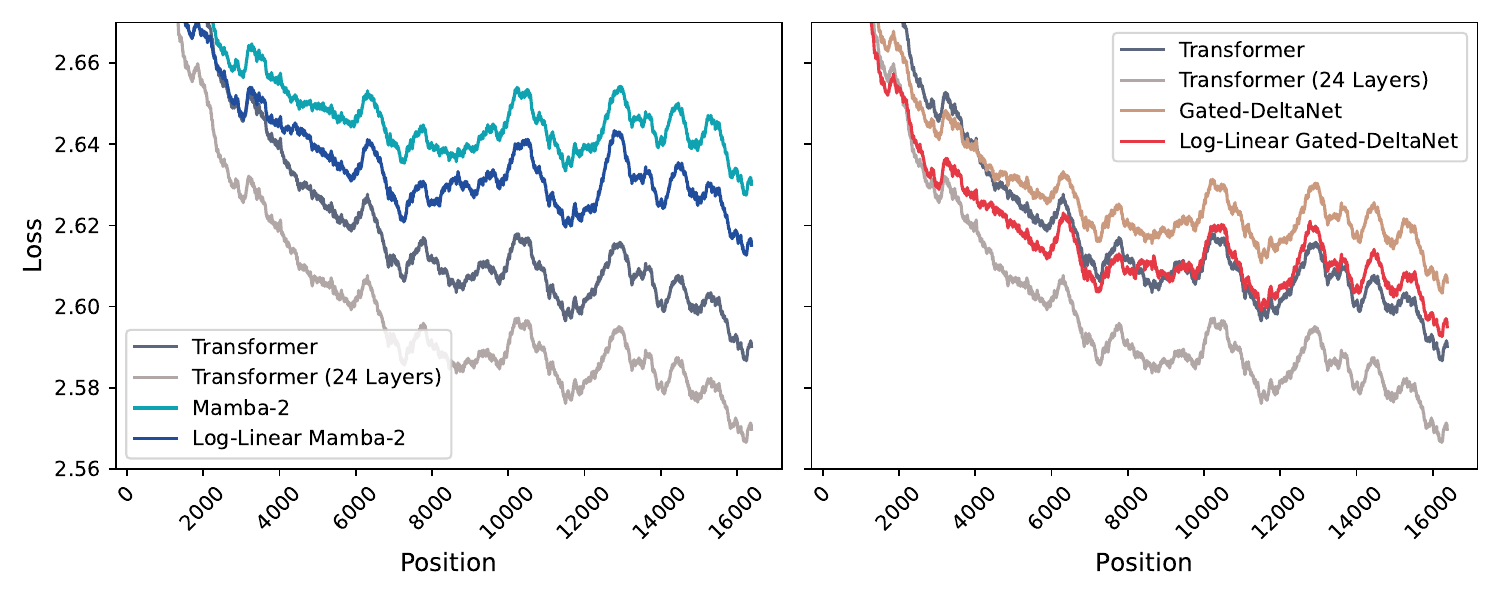}
\vspace{-3.9mm}
\captionof{figure}{Per-position loss on Book3 samples (about 39M tokens) with running average of window size $501$.
}
\label{fig:per-positional-loss}
\end{figure}
\begin{table*}[h]
\vspace{-5pt}
\centering
\small
\begin{tabular}{r|ccc|ccc|ccc}
\toprule
& \multicolumn{3}{c|}{\textbf{S-NIAH-1}}  & \multicolumn{3}{c|}{\textbf{S-NIAH-2}} & \multicolumn{3}{c}{\textbf{S-NIAH-3}}  \\
& \multicolumn{3}{c|}{(pass-key retrieval)}  & \multicolumn{3}{c|}{(number in haystack)} & \multicolumn{3}{c}{(uuid in haystack)}  \\
\textbf{Model}  & 4K  & 8K & 16K & 4K & 8K & 16K & 4K & 8K & 16K  \\
\midrule
Transformer          &  72.6 & 76.0 & 16.6 & 100.0 & 99.8 & 90.0 & 77.4 & 67.0 & 44.6 \\
w/ \emph{24 Layers}  &  92.4 & 78.4 & 89.8 & 100.0 & 100.0 & 99.6 & 84.0 & 63.6 & 36.4 \\
Mamba-2              & 90.4 & 56.8 & 21.6 & 72.4 & 28.0 & 18.6 & 4.0 & 3.6 & 0.8 \\
\rowcolor{black!10}
w/ \emph{Log-Linear} & 100.0 & 99.8 & 72.4 & 89.8 & 68.2 & 12.8 & 33.6 & 22.6 & 2.0 \\
Gated DeltaNet       & 100.0 & 100.0 & 100.0 & 95.8 & 46.8 & 5.0 & 66.2 & 14.6 & 6.0 \\
\rowcolor{black!10}
w/ \emph{Log-Linear} & 100.0 & 100.0 & 100.0 & 95.6 & 59.6 & 9.2 & 48.8 & 13.0 & 8.8 \\
\bottomrule
\addlinespace

& \multicolumn{3}{c|}{\textbf{MK-NIAH-1}} & \multicolumn{3}{c|}{\textbf{MQ-NIAH}}& \multicolumn{3}{c}{\textbf{MV-NIAH}} \\
& \multicolumn{3}{c|}{(multi-key line retrieval)} & \multicolumn{3}{c|}{(multi-query)} & \multicolumn{3}{c}{(multi-value)}  \\
\textbf{Model}  & 4K & 8K & 16K & 4K & 8K & 16K & 4K & 8K & 16K  \\
\midrule
Transformer          & 79.4 & 83.0 & 61.4 & 58.9 & 48.0 & 29.8 & 37.5 & 34.1 & 21.5 \\
w/ \emph{24 Layers}  & 62.6 & 83.2 & 75.2 & 54.6 & 46.0 & 34.5 & 48.4 & 45.4 & 32.3 \\
Mamba-2              & 27.2 & 18.6 & 13.6 & 28.7 & 19.4 & 1.3 & 27.9 & 14.8 & 4.4 \\
\rowcolor{black!10}
w/ \emph{Log-Linear} & 43.2 & 39.8 & 21.2 & 26.6 & 22.4 & 6.6 & 28.1 & 22.8 & 8.9 \\
Gated DeltaNet       & 23.0 & 21.2 & 5.2 & 21.6 & 16.9 & 7.2 & 16.2 & 14.5 & 7.0 \\
\rowcolor{black!10}
w/ \emph{Log-Linear} & 49.4 & 27.8 & 10.2 & 34.9 & 22.0 & 9.8 & 31.4 & 25.0 & 13.3 \\
\bottomrule
\end{tabular}
\centering
\vspace{-1mm}
\caption{
NIAH experiments with three single/multi-needle tasks.
}
\vspace{-6mm}
\label{tab:ruler}
\end{table*}

\paragraph{Per-position loss.}
Following~\citet{lin2025forgetting}, we report the model's loss at each token position to evaluate its ability to handle long contexts (Fig.~\ref{fig:per-positional-loss}). If the loss steadily decreases as the token position increases, it suggests the model is effectively using the full context. However, if the loss levels off after a certain point, it indicates the model struggles to make use of information that is too far back in the sequence. 
For this analysis, we use  39M tokens from Book-3.\footnote{\url{victor-wu/book3}} To improve visualization, we apply a running average with a window size of 501.
We observe that extending both Mamba-2 and Gated DeltaNet to their log-linear counterparts consistently reduces the (smoothed) loss across various positions, indicating improved long-range context utilization.
Log-Linear Gated DeltaNet also closely tracks the performance of the layer-matched Transformer, although a performance gap remains when compared to the parameter-matched Transformer.

\paragraph{Needle-In-A-Haystack.}
We use the Needle-In-A-Haystack (NIAH, Table~\ref{tab:ruler} and Fig.~\ref{fig:ruler}) benchmark from RULER~\citep{hsieh2024ruler}, where the model must retrieve a value (the “needle”) based on a key hidden in a long context (the “haystack”).
In the simpler single-needle tasks, the log-linear variant of Mamba-2 outperformed its linear counterpart on 8 out of 9 metrics. Gated DeltaNet, which already achieved perfect accuracy in several cases, saw improvements in 3 metrics, with 3 remaining unchanged. For the more challenging multi-needle tasks, {Log-Linear} Mamba-2 again improved in 8 out of 9 metrics, while {Log-Linear} Gated DeltaNet achieved improvements across all metrics.
\paragraph{Other tasks.} Due to space we show the results on 
the in-context retrieval benchmark \citep{arora2023zoology} and LongBench \citep{bai2023longbench} in the appendix.

\section{Discussion and Limitations}
While log-linear attention  improves upon linear attention in many cases, there are still quite a few tasks where it did not  improve upon the linear attention baselines. Due to compute resources we were unable to experiment with different parameterizations of the $\lambda$ terms (or hyperparameters in general),\footnote{We were only able to run our 700M-800M parameter language models just once due to compute constraints.} and it is possible that  optimal parameterization of $\lambda$ could lead to improved results. We also still observe a significant performance gap compared to Transformers across all benchmarks. 

The engineering complexity of log-linear attention is higher. Inter-chunk computations conceptually resemble multiple applications of linear attention primitives, but intra-chunk operations require bespoke implementations. These intra-chunk mechanisms are a primary factor behind the speed differences. Additionally, the backward pass is more intricate, as it requires (manually) computing the gradients not only for the standard attention components but also for the additional $\lambda$ terms.

The use of Fenwick-tree partitioning (\S\ref{subsec:fenwick-tree}) introduces an inductive bias: recent tokens are allocated more fine-grained memory, while distant tokens are compressed more aggressively. This design reflects a natural assumption rooted in hierarchical matrix which posits that interactions between distant elements can be approximated in low-rank form. While intuitive and inspired by physical phenomena, this inductive bias may not be optimal for all applications. Future work could explore extensions that enable more flexible structures while preserving computational efficiency.

Finally, in this work, we extended two existing linear-attention/SSM architectures to their log-linear counterparts, namely Mamba-2~\citep{dao2024transformers} and Gated DeltaNet~\citep{yang2024gated}. Several other promising architectures, including xLSTM~\citep{beckxlstm,beckxlstm7b} and MesaNet~\citep{von2023uncovering,von2025mesanet}, may likewise benefit from log-linear formulations. Developing and evaluating log-linear variants of these models is an exciting direction for future research.

\section{Related Work}
\vspace{-0.7mm}

\textbf{Structured matrices for deep learning architectures.}
Modern architectures combine token- and channel-mixing layers, both based on matrix multiplications. Recent work replaces dense layers with \emph{structured matrices}. For channel mixing, approaches include Butterfly \citep{dao2020learningfastalgorithmslinear}, Monarch matrices \citep{DBLP:conf/icml/DaoCSDPGLRRR22}, and more recently, Block Tensor-Train matrices \citep{Qiu2024ComputeBS}. Token mixing has been exemplified by the family of linear attention models \citep{katharopoulos2020transformers} and their various kernelizations \citep{xiong2021nystromformer}. \citet{dao2024transformers} generalize these approaches by extending low-rank structures to semiseparable matrices, enabling efficient recurrent inference and subsuming many recent recurrent models. Another line uses sparse patterns like sliding-window attention, alongside several hybrid methods~\citep{nguyen2021fmmformer,arora2025simplelinearattentionlanguage,munkhdalai2024leavecontextbehindefficient}.

\textbf{Log-linear complexity sequence modeling.}
Several prior efforts have focused on reducing the quadratic cost of attention to log-linear time complexity \citep{Kitaev2020Reformer, shi2023sequencemodelingmultiresolutionconvolutional, cunningham2024reparameterized, qin2023toeplitz, fu2023hungry,madaantreeformer,ye2019bp}.
Approaches such as LogSparse Transformer~\citep{li2019enhancing} and Informer~\citep{zhou2021informer} introduce sparse attention patterns to improve computational efficiency, particularly in time-series applications.
Reformer~\citep{Kitaev2020Reformer} employs locality-sensitive hashing (LSH) to efficiently cluster similar queries and keys. Multi-resolution attention~\citep{zeng2022multiresolutionanalysismra} adopts a hierarchical approach, progressively refining attention scores from coarse to fine granularity, while Fast Multipole Attention~\citep{kang2024fastmultipoleattentiondivideandconquer} adapts the classical fast multipole method to efficiently model long-range interactions.
A similar viewpoint connects log-linear attention to dilated convolution~\citep{van2016wavenet} through their hierarchical mixing structure. Dilated convolution extends convolution, which corresponds to Toeplitz matrices, whereas we operate primarily with semi-separable and hierarchical matrices.
In our work, we leverage the Fenwick tree data structure—a specialized binary indexed tree that enables efficient prefix sum calculations and updates in logarithmic time—to design an efficient attention layer during both training and decoding phases. 
While~\citet{zhu2021htransformer1dfastonedimensionalhierarchical} also employ hierarchical matrices for attention, their formulation is fully parallel and targeted at modest sequence lengths. In contrast, our approach adopts a chunkwise-parallel strategy with a custom Triton implementation optimized for long-sequence training. 
Concurrently,~\citet{yau2025sequential} propose a related architecture with $\mathcal{O}(\log T)$ memory, using a relaxed prefix-scan algorithm for state aggregation that accommodates arbitrary (potentially non-associative) functions.

\vspace{-1mm}
\section{Conclusion}
\vspace{-1mm}
We introduced Log-Linear Attention, a general framework that extends a broad class of linear attention and state-space models to their log-linear counterparts—models with logarithmically growing state size. This framework offers both theoretical insights and practical benefits, linking structured matrix theory with hardware-efficient computation. As a case study, we applied this approach to two recent architectures: Mamba-2 and Gated DeltaNet.

\section*{Acknowledgments}
We thank Tianyuan Zhang, Jyothish Pari, Adam Zweiger, and Yu Zhang for helpful discussion. This study was supported by the MIT-Google Program for Computing Innovation, MIT-IBM Watson AI Lab, and the AI2050 program at Schmidt Sciences (Grant G-25-67980). HG was supported by a Microsoft PhD Fellowship.
\bibliography{references}

\begin{thebibliography}{79}
\providecommand{\natexlab}[1]{#1}
\providecommand{\url}[1]{\texttt{#1}}
\expandafter\ifx\csname urlstyle\endcsname\relax
  \providecommand{\doi}[1]{doi: #1}\else
  \providecommand{\doi}{doi: \begingroup \urlstyle{rm}\Url}\fi

\bibitem[Arora et~al.(2023)Arora, Eyuboglu, Timalsina, Johnson, Poli, Zou, Rudra, and R{\'e}]{arora2023zoology}
Simran Arora, Sabri Eyuboglu, Aman Timalsina, Isys Johnson, Michael Poli, James Zou, Atri Rudra, and Christopher R{\'e}.
\newblock Zoology: Measuring and improving recall in efficient language models.
\newblock \emph{arXiv preprint arXiv:2312.04927}, 2023.

\bibitem[Arora et~al.(2024)Arora, Eyuboglu, Zhang, Timalsina, Alberti, Zinsley, Zou, Rudra, and R{\'e}]{arora2024simple}
Simran Arora, Sabri Eyuboglu, Michael Zhang, Aman Timalsina, Silas Alberti, Dylan Zinsley, James Zou, Atri Rudra, and Christopher R{\'e}.
\newblock Simple linear attention language models balance the recall-throughput tradeoff.
\newblock In \emph{Proceedings of ICML}, 2024.

\bibitem[Arora et~al.(2025)Arora, Eyuboglu, Zhang, Timalsina, Alberti, Zinsley, Zou, Rudra, and Ré]{arora2025simplelinearattentionlanguage}
Simran Arora, Sabri Eyuboglu, Michael Zhang, Aman Timalsina, Silas Alberti, Dylan Zinsley, James Zou, Atri Rudra, and Christopher Ré.
\newblock Simple linear attention language models balance the recall-throughput tradeoff, 2025.
\newblock URL \url{https://arxiv.org/abs/2402.18668}.

\bibitem[Bahdanau et~al.(2014)Bahdanau, Cho, and Bengio]{bahdanau2014neural}
Dzmitry Bahdanau, Kyunghyun Cho, and Yoshua Bengio.
\newblock Neural machine translation by jointly learning to align and translate.
\newblock In \emph{Proceedings of ICLR}, 2014.

\bibitem[Bai et~al.(2023)Bai, Lv, Zhang, Lyu, Tang, Huang, Du, Liu, Zeng, Hou, et~al.]{bai2023longbench}
Yushi Bai, Xin Lv, Jiajie Zhang, Hongchang Lyu, Jiankai Tang, Zhidian Huang, Zhengxiao Du, Xiao Liu, Aohan Zeng, Lei Hou, et~al.
\newblock Longbench: A bilingual, multitask benchmark for long context understanding.
\newblock \emph{arXiv preprint arXiv:2308.14508}, 2023.

\bibitem[Beck et~al.(2024)Beck, P{\"o}ppel, Spanring, Auer, Prudnikova, Kopp, Klambauer, Brandstetter, and Hochreiter]{beckxlstm}
Maximilian Beck, Korbinian P{\"o}ppel, Markus Spanring, Andreas Auer, Oleksandra Prudnikova, Michael~K Kopp, G{\"u}nter Klambauer, Johannes Brandstetter, and Sepp Hochreiter.
\newblock x{LSTM}: Extended long short-term memory.
\newblock In \emph{The Thirty-eighth Annual Conference on Neural Information Processing Systems}, 2024.
\newblock URL \url{https://openreview.net/forum?id=ARAxPPIAhq}.

\bibitem[Beck et~al.(2025{\natexlab{a}})Beck, P{\"o}ppel, Lippe, and Hochreiter]{beck2025tiled}
Maximilian Beck, Korbinian P{\"o}ppel, Phillip Lippe, and Sepp Hochreiter.
\newblock Tiled flash linear attention: More efficient linear rnn and xlstm kernels.
\newblock \emph{arXiv preprint arXiv:2503.14376}, 2025{\natexlab{a}}.

\bibitem[Beck et~al.(2025{\natexlab{b}})Beck, P{\"o}ppel, Lippe, Kurle, Blies, Klambauer, B{\"o}ck, and Hochreiter]{beckxlstm7b}
Maximilian Beck, Korbinian P{\"o}ppel, Phillip Lippe, Richard Kurle, Patrick~M Blies, G{\"u}nter Klambauer, Sebastian B{\"o}ck, and Sepp Hochreiter.
\newblock x{LSTM} 7b: A recurrent {LLM} for fast and efficient inference.
\newblock In \emph{Forty-second International Conference on Machine Learning}, 2025{\natexlab{b}}.
\newblock URL \url{https://openreview.net/forum?id=LV3DpKD08B}.

\bibitem[Bischof \& Loan(1985)Bischof and Loan]{bischof_wy_1985}
Christian~H. Bischof and Charles~Van Loan.
\newblock The {WY} representation for products of householder matrices.
\newblock In \emph{{SIAM} {Conference} on {Parallel} {Processing} for {Scientific} {Computing}}, 1985.
\newblock URL \url{https://api.semanticscholar.org/CorpusID:36094006}.

\bibitem[Blelloch(1990)]{blelloch1990prefix}
Guy~E Blelloch.
\newblock Prefix sums and their applications.
\newblock 1990.

\bibitem[Buckman et~al.()Buckman, Gelada, and Zhang]{buckman2024}
Jacob Buckman, Carles Gelada, and Sean Zhang.
\newblock Symmetric {Power} {Transformers}.

\bibitem[Cunningham et~al.(2024)Cunningham, Giannone, Zhang, and Deisenroth]{cunningham2024reparameterized}
Harry~Jake Cunningham, Giorgio Giannone, Mingtian Zhang, and Marc~Peter Deisenroth.
\newblock Reparameterized multi-resolution convolutions for long sequence modelling.
\newblock In \emph{The Thirty-eighth Annual Conference on Neural Information Processing Systems}, 2024.
\newblock URL \url{https://openreview.net/forum?id=RwgNbIpCpk}.

\bibitem[Dao(2024)]{dao2023flashattention2}
Tri Dao.
\newblock Flash{A}ttention-2: Faster attention with better parallelism and work partitioning.
\newblock In \emph{Proceedings of ICLR}, 2024.

\bibitem[Dao \& Gu(2024)Dao and Gu]{dao2024transformers}
Tri Dao and Albert Gu.
\newblock Transformers are {SSM}s: Generalized models and efficient algorithms through structured state space duality.
\newblock In \emph{Proceedings of ICML}, 2024.

\bibitem[Dao et~al.(2020)Dao, Gu, Eichhorn, Rudra, and Ré]{dao2020learningfastalgorithmslinear}
Tri Dao, Albert Gu, Matthew Eichhorn, Atri Rudra, and Christopher Ré.
\newblock Learning fast algorithms for linear transforms using butterfly factorizations, 2020.
\newblock URL \url{https://arxiv.org/abs/1903.05895}.

\bibitem[Dao et~al.(2022{\natexlab{a}})Dao, Chen, Sohoni, Desai, Poli, Grogan, Liu, Rao, Rudra, and R{\'{e}}]{DBLP:conf/icml/DaoCSDPGLRRR22}
Tri Dao, Beidi Chen, Nimit~Sharad Sohoni, Arjun~D. Desai, Michael Poli, Jessica Grogan, Alexander Liu, Aniruddh Rao, Atri Rudra, and Christopher R{\'{e}}.
\newblock Monarch: Expressive structured matrices for efficient and accurate training.
\newblock In Kamalika Chaudhuri, Stefanie Jegelka, Le~Song, Csaba Szepesv{\'{a}}ri, Gang Niu, and Sivan Sabato (eds.), \emph{International Conference on Machine Learning, {ICML} 2022, 17-23 July 2022, Baltimore, Maryland, {USA}}, volume 162 of \emph{Proceedings of Machine Learning Research}, pp.\  4690--4721. {PMLR}, 2022{\natexlab{a}}.
\newblock URL \url{https://proceedings.mlr.press/v162/dao22a.html}.

\bibitem[Dao et~al.(2022{\natexlab{b}})Dao, Fu, Ermon, Rudra, and R{\'e}]{dao2022flashattention}
Tri Dao, Daniel~Y. Fu, Stefano Ermon, Atri Rudra, and Christopher R{\'e}.
\newblock Flash{A}ttention: Fast and memory-efficient exact attention with {IO}-awareness.
\newblock In \emph{Proceedings of NeurIPS}, 2022{\natexlab{b}}.

\bibitem[Fenwick(1994)]{Fenwick1994AND}
Peter~M. Fenwick.
\newblock A new data structure for cumulative frequency tables.
\newblock \emph{Software: Practice and Experience}, 24, 1994.
\newblock URL \url{https://api.semanticscholar.org/CorpusID:7519761}.

\bibitem[Fu et~al.(2023)Fu, Dao, Saab, Thomas, Rudra, and Re]{fu2023hungry}
Daniel~Y Fu, Tri Dao, Khaled~Kamal Saab, Armin~W Thomas, Atri Rudra, and Christopher Re.
\newblock Hungry hungry hippos: Towards language modeling with state space models.
\newblock In \emph{The Eleventh International Conference on Learning Representations}, 2023.
\newblock URL \url{https://openreview.net/forum?id=COZDy0WYGg}.

\bibitem[Grazzi et~al.(2025)Grazzi, Siems, Franke, Zela, Hutter, and Pontil]{grazzi2025unlocking}
Riccardo Grazzi, Julien Siems, J{\"o}rg~K.H. Franke, Arber Zela, Frank Hutter, and Massimiliano Pontil.
\newblock Unlocking state-tracking in linear {RNN}s through negative eigenvalues.
\newblock In \emph{Proceedings of ICLR}, 2025.

\bibitem[Gu \& Dao(2024)Gu and Dao]{Gu2023MambaLS}
Albert Gu and Tri Dao.
\newblock Mamba: Linear-time sequence modeling with selective state spaces.
\newblock In \emph{Proceedings of CoLM}, 2024.

\bibitem[Gu et~al.(2022)Gu, Goel, and R{\'e}]{gu2022efficiently}
Albert Gu, Karan Goel, and Christopher R{\'e}.
\newblock Efficiently modeling long sequences with structured state spaces.
\newblock In \emph{Proceedings of ICLR}, 2022.

\bibitem[Hackbusch et~al.(2004)Hackbusch, Khoromskij, and Kriemann]{hackbusch2004hierarchical}
Wolfgang Hackbusch, Boris~N Khoromskij, and Ronald Kriemann.
\newblock Hierarchical matrices based on a weak admissibility criterion.
\newblock \emph{Computing}, 73:\penalty0 207--243, 2004.

\bibitem[Hsieh et~al.(2024)Hsieh, Sun, Kriman, Acharya, Rekesh, Jia, Zhang, and Ginsburg]{hsieh2024ruler}
Cheng-Ping Hsieh, Simeng Sun, Samuel Kriman, Shantanu Acharya, Dima Rekesh, Fei Jia, Yang Zhang, and Boris Ginsburg.
\newblock Ruler: What's the real context size of your long-context language models?
\newblock \emph{arXiv preprint arXiv:2404.06654}, 2024.

\bibitem[Hua et~al.(2022)Hua, Dai, Liu, and Le]{hua2022transformer}
Weizhe Hua, Zihang Dai, Hanxiao Liu, and Quoc Le.
\newblock Transformer quality in linear time.
\newblock In \emph{International conference on machine learning}, pp.\  9099--9117. PMLR, 2022.

\bibitem[Joffrain et~al.(2006)Joffrain, Low, Quintana-Ort{\'i}, van~de Geijn, and Zee]{Joffrain2006AccumulatingHT}
Thierry Joffrain, Tze~Meng Low, Enrique~S. Quintana-Ort{\'i}, Robert~A. van~de Geijn, and Field G.~Van Zee.
\newblock Accumulating householder transformations, revisited.
\newblock \emph{ACM Trans. Math. Softw.}, 32:\penalty0 169--179, 2006.
\newblock URL \url{https://api.semanticscholar.org/CorpusID:15723171}.

\bibitem[Kacham et~al.(2023)Kacham, Mirrokni, and Zhong]{kacham2023polysketchformer}
Praneeth Kacham, Vahab Mirrokni, and Peilin Zhong.
\newblock Polysketchformer: Fast transformers via sketching polynomial kernels.
\newblock \emph{arXiv preprint arXiv:2310.01655}, 2023.

\bibitem[Kang et~al.(2024)Kang, Tran, and Sterck]{kang2024fastmultipoleattentiondivideandconquer}
Yanming Kang, Giang Tran, and Hans~De Sterck.
\newblock Fast multipole attention: A divide-and-conquer attention mechanism for long sequences, 2024.
\newblock URL \url{https://arxiv.org/abs/2310.11960}.

\bibitem[Kasai et~al.(2021)Kasai, Peng, Zhang, Yogatama, Ilharco, Pappas, Mao, Chen, and Smith]{kasai2021finetuning}
Jungo Kasai, Hao Peng, Yizhe Zhang, Dani Yogatama, Gabriel Ilharco, Nikolaos Pappas, Yi~Mao, Weizhu Chen, and Noah~A Smith.
\newblock Finetuning pretrained transformers into rnns.
\newblock In \emph{Proceedings of EMNLP}, 2021.

\bibitem[Katharopoulos et~al.(2020)Katharopoulos, Vyas, Pappas, and Fleuret]{katharopoulos2020transformers}
Angelos Katharopoulos, Apoorv Vyas, Nikolaos Pappas, and Fran{\c{c}}ois Fleuret.
\newblock Transformers are rnns: Fast autoregressive transformers with linear attention.
\newblock In \emph{Proceedings of ICML}, 2020.

\bibitem[Katsch(2023)]{katsch2023gateloop}
Tobias Katsch.
\newblock Gateloop: Fully data-controlled linear recurrence for sequence modeling.
\newblock \emph{arXiv preprint arXiv:2311.01927}, 2023.

\bibitem[Kitaev et~al.(2020)Kitaev, Kaiser, and Levskaya]{Kitaev2020Reformer}
Nikita Kitaev, {\L}ukasz Kaiser, and Anselm Levskaya.
\newblock Reformer: The efficient transformer.
\newblock In \emph{Proceedings of ICLR}, 2020.

\bibitem[Kressner et~al.(2019)Kressner, Massei, and Robol]{kressner2019low}
Daniel Kressner, Stefano Massei, and Leonardo Robol.
\newblock Low-rank updates and a divide-and-conquer method for linear matrix equations.
\newblock \emph{SIAM Journal on Scientific Computing}, 41\penalty0 (2):\penalty0 A848--A876, 2019.

\bibitem[Kwon et~al.(2023)Kwon, Li, Zhuang, Sheng, Zheng, Yu, Gonzalez, Zhang, and Stoica]{kwon2023efficient}
Woosuk Kwon, Zhuohan Li, Siyuan Zhuang, Ying Sheng, Lianmin Zheng, Cody~Hao Yu, Joseph Gonzalez, Hao Zhang, and Ion Stoica.
\newblock Efficient memory management for large language model serving with pagedattention.
\newblock In \emph{Proceedings of SOSP}, 2023.

\bibitem[Li et~al.(2019)Li, Jin, Xuan, Zhou, Chen, Wang, and Yan]{li2019enhancing}
Shiyang Li, Xiaoyong Jin, Yao Xuan, Xiyou Zhou, Wenhu Chen, Yu-Xiang Wang, and Xifeng Yan.
\newblock Enhancing the locality and breaking the memory bottleneck of transformer on time series forecasting.
\newblock \emph{Advances in neural information processing systems}, 32, 2019.

\bibitem[Lin et~al.(2025)Lin, Nikishin, He, and Courville]{lin2025forgetting}
Zhixuan Lin, Evgenii Nikishin, Xu~He, and Aaron Courville.
\newblock Forgetting transformer: Softmax attention with a forget gate.
\newblock In \emph{The Thirteenth International Conference on Learning Representations}, 2025.
\newblock URL \url{https://openreview.net/forum?id=q2Lnyegkr8}.

\bibitem[Liu et~al.(2024)Liu, Zaharia, and Abbeel]{liu2023ring}
Hao Liu, Matei Zaharia, and Pieter Abbeel.
\newblock Ring attention with blockwise transformers for near-infinite context.
\newblock In \emph{Proceedings of ICLR}, 2024.

\bibitem[Madaan et~al.()Madaan, Bhojanapalli, Jain, and Jain]{madaantreeformer}
Lovish Madaan, Srinadh Bhojanapalli, Himanshu Jain, and Prateek Jain.
\newblock Treeformer: Dense gradient trees for efficient attention computation.
\newblock In \emph{The Eleventh International Conference on Learning Representations}.

\bibitem[Mao()]{mao_fine-tuning_2022}
Huanru~Henry Mao.
\newblock Fine-{Tuning} {Pre}-trained {Transformers} into {Decaying} {Fast} {Weights}.
\newblock In \emph{Proceedings of EMNLP}, pp.\  10236--10242.

\bibitem[Massaroli et~al.(2023)Massaroli, Poli, Fu, Kumbong, Parnichkun, Romero, Timalsina, McIntyre, Chen, Rudra, Zhang, Re, Ermon, and Bengio]{massaroli2023laughing}
Stefano Massaroli, Michael Poli, Daniel~Y Fu, Hermann Kumbong, Rom~Nishijima Parnichkun, David~W. Romero, Aman Timalsina, Quinn McIntyre, Beidi Chen, Atri Rudra, Ce~Zhang, Christopher Re, Stefano Ermon, and Yoshua Bengio.
\newblock Laughing hyena distillery: Extracting compact recurrences from convolutions.
\newblock In \emph{Thirty-seventh Conference on Neural Information Processing Systems}, 2023.
\newblock URL \url{https://openreview.net/forum?id=OWELckerm6}.

\bibitem[Massei et~al.(2020)Massei, Robol, and Kressner]{massei2020hm}
Stefano Massei, Leonardo Robol, and Daniel Kressner.
\newblock hm-toolbox: Matlab software for hodlr and hss matrices.
\newblock \emph{SIAM Journal on Scientific Computing}, 42\penalty0 (2):\penalty0 C43--C68, 2020.

\bibitem[Merrill et~al.(2024)Merrill, Petty, and Sabharwal]{merrill2024illusion}
William Merrill, Jackson Petty, and Ashish Sabharwal.
\newblock The illusion of state in state-space models.
\newblock In \emph{Proceedings of ICML}, 2024.

\bibitem[Munkhdalai et~al.(2024)Munkhdalai, Faruqui, and Gopal]{munkhdalai2024leavecontextbehindefficient}
Tsendsuren Munkhdalai, Manaal Faruqui, and Siddharth Gopal.
\newblock Leave no context behind: Efficient infinite context transformers with infini-attention, 2024.
\newblock URL \url{https://arxiv.org/abs/2404.07143}.

\bibitem[Nguyen et~al.(2021)Nguyen, Suliafu, Osher, Chen, and Wang]{nguyen2021fmmformer}
Tan Nguyen, Vai Suliafu, Stanley Osher, Long Chen, and Bao Wang.
\newblock Fmmformer: Efficient and flexible transformer via decomposed near-field and far-field attention.
\newblock In \emph{Proceedings of NeurIPS}, 2021.

\bibitem[Oncescu et~al.(2025)Oncescu, Purandare, Idreos, and Kakade]{oncescu2025flash}
Costin-Andrei Oncescu, Sanket Purandare, Stratos Idreos, and Sham~M. Kakade.
\newblock Flash inference: Near linear time inference for long convolution sequence models and beyond.
\newblock In \emph{The Thirteenth International Conference on Learning Representations}, 2025.
\newblock URL \url{https://openreview.net/forum?id=cZWCjan02B}.

\bibitem[Peng et~al.(2024)Peng, Goldstein, Anthony, Albalak, Alcaide, Biderman, Cheah, Ferdinan, Hou, Kazienko, et~al.]{peng2024eagle}
Bo~Peng, Daniel Goldstein, Quentin Anthony, Alon Albalak, Eric Alcaide, Stella Biderman, Eugene Cheah, Teddy Ferdinan, Haowen Hou, Przemys{\l}aw Kazienko, et~al.
\newblock Eagle and finch: Rwkv with matrix-valued states and dynamic recurrence.
\newblock \emph{arXiv preprint arXiv:2404.05892}, 3, 2024.

\bibitem[Peng et~al.(2025)Peng, Zhang, Goldstein, Alcaide, Du, Hou, Lin, Liu, Lu, Merrill, et~al.]{peng2025rwkv}
Bo~Peng, Ruichong Zhang, Daniel Goldstein, Eric Alcaide, Xingjian Du, Haowen Hou, Jiaju Lin, Jiaxing Liu, Janna Lu, William Merrill, et~al.
\newblock Rwkv-7" goose" with expressive dynamic state evolution.
\newblock \emph{arXiv preprint arXiv:2503.14456}, 2025.

\bibitem[Peng et~al.(2021)Peng, Pappas, Yogatama, Schwartz, Smith, and Kong]{peng_random_2021}
Hao Peng, Nikolaos Pappas, Dani Yogatama, Roy Schwartz, Noah~A. Smith, and Lingpeng Kong.
\newblock In \emph{Proceedings of ICLR}, 2021.

\bibitem[Poli et~al.(2023)Poli, Massaroli, Nguyen, Fu, Dao, Baccus, Bengio, Ermon, and R{\'{e}}]{hyena}
Michael Poli, Stefano Massaroli, Eric Nguyen, Daniel~Y. Fu, Tri Dao, Stephen Baccus, Yoshua Bengio, Stefano Ermon, and Christopher R{\'{e}}.
\newblock Hyena hierarchy: Towards larger convolutional language models.
\newblock In Andreas Krause, Emma Brunskill, Kyunghyun Cho, Barbara Engelhardt, Sivan Sabato, and Jonathan Scarlett (eds.), \emph{International Conference on Machine Learning, {ICML} 2023, 23-29 July 2023, Honolulu, Hawaii, {USA}}, volume 202 of \emph{Proceedings of Machine Learning Research}, pp.\  28043--28078. {PMLR}, 2023.
\newblock URL \url{https://proceedings.mlr.press/v202/poli23a.html}.

\bibitem[Qin \& Zhong(2023)Qin and Zhong]{qin-zhong-2023-accelerating}
Zhen Qin and Yiran Zhong.
\newblock Accelerating toeplitz neural network with constant-time inference complexity.
\newblock In Houda Bouamor, Juan Pino, and Kalika Bali (eds.), \emph{Proceedings of the 2023 Conference on Empirical Methods in Natural Language Processing}, pp.\  12206--12215, Singapore, December 2023. Association for Computational Linguistics.
\newblock \doi{10.18653/v1/2023.emnlp-main.750}.
\newblock URL \url{https://aclanthology.org/2023.emnlp-main.750/}.

\bibitem[Qin et~al.(2022)Qin, Sun, Deng, Li, Wei, Lv, Yan, Kong, and Zhong]{qin2022cosformer}
Zhen Qin, Weixuan Sun, Hui Deng, Dongxu Li, Yunshen Wei, Baohong Lv, Junjie Yan, Lingpeng Kong, and Yiran Zhong.
\newblock cosformer: Rethinking softmax in attention.
\newblock In \emph{Proceedings of ICLR}, 2022.

\bibitem[Qin et~al.(2023)Qin, Han, Sun, He, Li, Li, Dai, Kong, and Zhong]{qin2023toeplitz}
Zhen Qin, Xiaodong Han, Weixuan Sun, Bowen He, Dong Li, Dongxu Li, Yuchao Dai, Lingpeng Kong, and Yiran Zhong.
\newblock Toeplitz neural network for sequence modeling.
\newblock In \emph{Proceedings of ICLR}, 2023.

\bibitem[Qin et~al.(2024{\natexlab{a}})Qin, Sun, Li, Shen, Sun, and Zhong]{qin2024lightning}
Zhen Qin, Weigao Sun, Dong Li, Xuyang Shen, Weixuan Sun, and Yiran Zhong.
\newblock Lightning attention-2: A free lunch for handling unlimited sequence lengths in large language models.
\newblock \emph{arXiv preprint arXiv:2401.04658}, 2024{\natexlab{a}}.

\bibitem[Qin et~al.(2024{\natexlab{b}})Qin, Yang, Sun, Shen, Li, Sun, and Zhong]{qin_hgrn2_2024}
Zhen Qin, Songlin Yang, Weixuan Sun, Xuyang Shen, Dong Li, Weigao Sun, and Yiran Zhong.
\newblock {HGRN2}: {Gated} {Linear} {RNNs} with {State} {Expansion}.
\newblock In \emph{Proceedings of CoLM}, 2024{\natexlab{b}}.

\bibitem[Qiu et~al.(2024)Qiu, Potapczynski, Finzi, Goldblum, and Wilson]{Qiu2024ComputeBS}
Shikai Qiu, Andres Potapczynski, Marc Finzi, Micah Goldblum, and Andrew~Gordon Wilson.
\newblock Compute better spent: Replacing dense layers with structured matrices.
\newblock \emph{ArXiv}, abs/2406.06248, 2024.
\newblock URL \url{https://api.semanticscholar.org/CorpusID:270371652}.

\bibitem[Ryabko(1992)]{Ryabko1992AFO}
B.~Ya. Ryabko.
\newblock A fast on-line adaptive code.
\newblock \emph{IEEE Trans. Inf. Theory}, 38:\penalty0 1400--1404, 1992.
\newblock URL \url{https://api.semanticscholar.org/CorpusID:206392294}.

\bibitem[Schlag et~al.(2021)Schlag, Irie, and Schmidhuber]{schlag_linear_2021}
Imanol Schlag, Kazuki Irie, and J{\"u}rgen Schmidhuber.
\newblock Linear {Transformers} {Are} {Secretly} {Fast} {Weight} {Programmers}.
\newblock In \emph{Proceedings of ICML}, 2021.

\bibitem[Schmidhuber(1992)]{schmidhuber1992learning}
J{\"u}rgen Schmidhuber.
\newblock Learning to control fast-weight memories: An alternative to dynamic recurrent networks.
\newblock \emph{Neural Computation}, 4\penalty0 (1):\penalty0 131--139, 1992.

\bibitem[Shah et~al.(2024)Shah, Bikshandi, Zhang, Thakkar, Ramani, and Dao]{shah2024flashattention}
Jay Shah, Ganesh Bikshandi, Ying Zhang, Vijay Thakkar, Pradeep Ramani, and Tri Dao.
\newblock Flash{A}ttention-3: Fast and accurate attention with asynchrony and low-precision.
\newblock In \emph{Proceedings of NeurIPS}, 2024.

\bibitem[Shi et~al.(2023)Shi, Wang, and Fox]{shi2023sequencemodelingmultiresolutionconvolutional}
Jiaxin Shi, Ke~Alexander Wang, and Emily~B. Fox.
\newblock Sequence modeling with multiresolution convolutional memory, 2023.
\newblock URL \url{https://arxiv.org/abs/2305.01638}.

\bibitem[Siems et~al.(2025)Siems, Carstensen, Zela, Hutter, Pontil, and Grazzi]{siems2025deltaproduct}
Julien Siems, Timur Carstensen, Arber Zela, Frank Hutter, Massimiliano Pontil, and Riccardo Grazzi.
\newblock Deltaproduct: Improving state-tracking in linear rnns via householder products.
\newblock \emph{arXiv preprint arXiv:2502.10297}, 2025.

\bibitem[Sun et~al.(2023)Sun, Dong, Huang, Ma, Xia, Xue, Wang, and Wei]{sun2023retentive}
Yutao Sun, Li~Dong, Shaohan Huang, Shuming Ma, Yuqing Xia, Jilong Xue, Jianyong Wang, and Furu Wei.
\newblock Retentive network: A successor to transformer for large language models.
\newblock \emph{arXiv preprint arXiv:2307.08621}, 2023.

\bibitem[Tillet et~al.(2019)Tillet, Kung, and Cox]{tillet2019triton}
Philippe Tillet, Hsiang-Tsung Kung, and David Cox.
\newblock Triton: an intermediate language and compiler for tiled neural network computations.
\newblock In \emph{Proceedings of the 3rd ACM SIGPLAN International Workshop on Machine Learning and Programming Languages}, pp.\  10--19, 2019.

\bibitem[Van Den~Oord et~al.(2016)Van Den~Oord, Dieleman, Zen, Simonyan, Vinyals, Graves, Kalchbrenner, Senior, Kavukcuoglu, et~al.]{van2016wavenet}
Aaron Van Den~Oord, Sander Dieleman, Heiga Zen, Karen Simonyan, Oriol Vinyals, Alex Graves, Nal Kalchbrenner, Andrew Senior, Koray Kavukcuoglu, et~al.
\newblock Wavenet: A generative model for raw audio.
\newblock \emph{arXiv preprint arXiv:1609.03499}, 12:\penalty0 1, 2016.

\bibitem[Vaswani et~al.(2017)Vaswani, Shazeer, Parmar, Uszkoreit, Jones, Gomez, Kaiser, and Polosukhin]{vaswani2017attention}
Ashish Vaswani, Noam Shazeer, Niki Parmar, Jakob Uszkoreit, Llion Jones, Aidan~N Gomez, {\L}ukasz Kaiser, and Illia Polosukhin.
\newblock Attention is all you need.
\newblock In \emph{Proceedings of NeurIPS}, 2017.

\bibitem[Von~Oswald et~al.(2023)Von~Oswald, Schlegel, Meulemans, Kobayashi, Niklasson, Zucchet, Scherrer, Miller, Sandler, Vladymyrov, et~al.]{von2023uncovering}
Johannes Von~Oswald, Maximilian Schlegel, Alexander Meulemans, Seijin Kobayashi, Eyvind Niklasson, Nicolas Zucchet, Nino Scherrer, Nolan Miller, Mark Sandler, Max Vladymyrov, et~al.
\newblock Uncovering mesa-optimization algorithms in transformers.
\newblock \emph{arXiv preprint arXiv:2309.05858}, 2023.

\bibitem[von Oswald et~al.(2025)von Oswald, Scherrer, Kobayashi, Versari, Yang, Schlegel, Maile, Schimpf, Sieberling, Meulemans, et~al.]{von2025mesanet}
Johannes von Oswald, Nino Scherrer, Seijin Kobayashi, Luca Versari, Songlin Yang, Maximilian Schlegel, Kaitlin Maile, Yanick Schimpf, Oliver Sieberling, Alexander Meulemans, et~al.
\newblock Mesanet: Sequence modeling by locally optimal test-time training.
\newblock \emph{arXiv preprint arXiv:2506.05233}, 2025.

\bibitem[Widrow et~al.(1960)Widrow, Hoff, et~al.]{widrow_adaptive_1988}
Bernard Widrow, Marcian~E Hoff, et~al.
\newblock Adaptive switching circuits.
\newblock In \emph{IRE WESCON convention record}, volume~4, pp.\  96--104. New York, 1960.

\bibitem[Xiong et~al.(2021)Xiong, Zeng, Chakraborty, Tan, Fung, Li, and Singh]{xiong2021nystromformer}
Yunyang Xiong, Zhanpeng Zeng, Rudrasis Chakraborty, Mingxing Tan, Glenn Fung, Yin Li, and Vikas Singh.
\newblock Nystr{\"o}mformer: A nystr{\"o}m-based algorithm for approximating self-attention.
\newblock In \emph{Proceedings of AAAI}, 2021.

\bibitem[Yang \& Zhang(2024)Yang and Zhang]{yang2024fla}
Songlin Yang and Yu~Zhang.
\newblock Fla: A triton-based library for hardware-efficient implementations of linear attention mechanism, January 2024.
\newblock URL \url{https://github.com/fla-org/flash-linear-attention}.

\bibitem[Yang et~al.(2023)Yang, Wang, Shen, Panda, and Kim]{yang2023gated}
Songlin Yang, Bailin Wang, Yikang Shen, Rameswar Panda, and Yoon Kim.
\newblock Gated linear attention transformers with hardware-efficient training.
\newblock \emph{arXiv preprint arXiv:2312.06635}, 2023.

\bibitem[Yang et~al.(2024{\natexlab{a}})Yang, Kautz, and Hatamizadeh]{yang2024gated}
Songlin Yang, Jan Kautz, and Ali Hatamizadeh.
\newblock Gated delta networks: Improving mamba2 with delta rule.
\newblock In \emph{Proceedings of ICLR}, 2024{\natexlab{a}}.

\bibitem[Yang et~al.(2024{\natexlab{b}})Yang, Wang, Zhang, Shen, and Kim]{yang2024parallelizing}
Songlin Yang, Bailin Wang, Yu~Zhang, Yikang Shen, and Yoon Kim.
\newblock Parallelizing linear transformers with the delta rule over sequence length.
\newblock In \emph{Proceedings of NeurIPS}, 2024{\natexlab{b}}.

\bibitem[Yau et~al.(2025)Yau, Gupta, Engelmayer, Irie, Jegelka, and Andreas]{yau2025sequential}
Morris Yau, Sharut Gupta, Valerie Engelmayer, Kazuki Irie, Stefanie Jegelka, and Jacob Andreas.
\newblock Sequential-parallel duality in prefix scannable models.
\newblock \emph{arXiv preprint arXiv:2506.10918}, 2025.

\bibitem[Ye et~al.(2019)Ye, Guo, Gan, Qiu, and Zhang]{ye2019bp}
Zihao Ye, Qipeng Guo, Quan Gan, Xipeng Qiu, and Zheng Zhang.
\newblock Bp-transformer: Modelling long-range context via binary partitioning.
\newblock \emph{arXiv preprint arXiv:1911.04070}, 2019.

\bibitem[Zeng et~al.(2022)Zeng, Pal, Kline, Fung, and Singh]{zeng2022multiresolutionanalysismra}
Zhanpeng Zeng, Sourav Pal, Jeffery Kline, Glenn~M Fung, and Vikas Singh.
\newblock Multi resolution analysis (mra) for approximate self-attention, 2022.
\newblock URL \url{https://arxiv.org/abs/2207.10284}.

\bibitem[Zhang \& Yang(2025)Zhang and Yang]{yang2025flame}
Yu~Zhang and Songlin Yang.
\newblock Flame: Flash language modeling made easy, January 2025.
\newblock URL \url{https://github.com/fla-org/flame}.

\bibitem[Zhou et~al.(2021)Zhou, Zhang, Peng, Zhang, Li, Xiong, and Zhang]{zhou2021informer}
Haoyi Zhou, Shanghang Zhang, Jieqi Peng, Shuai Zhang, Jianxin Li, Hui Xiong, and Wancai Zhang.
\newblock Informer: Beyond efficient transformer for long sequence time-series forecasting.
\newblock In \emph{Proceedings of AAAI}, 2021.

\bibitem[Zhu \& Soricut(2021)Zhu and Soricut]{zhu2021htransformer1dfastonedimensionalhierarchical}
Zhenhai Zhu and Radu Soricut.
\newblock H-transformer-1d: Fast one-dimensional hierarchical attention for sequences, 2021.
\newblock URL \url{https://arxiv.org/abs/2107.11906}.

\end{thebibliography}
\bibliographystyle{iclr2026_conference}

\appendix
\newpage

\section{Generalizing Log-Linear Attention to More Expressive Linear RNNs}
\label{appendix-subsec:4d-tensor-view}

The main paper adopts the following unified view of efficient attention (Eq.~\ref{eq:efficient-attention-as-structured-matrices}):
\begin{equation*}
\mP = \mA \odot \mM, \quad \mO = \mP \mV,
\end{equation*}
This formulation reveals that the key difference between linear and log-linear attention lies in the structure of the mask matrix $\mM \in \mathbb{R}^{T \times T}$. Variations among linear attention models—such as Mamba-2 and Gated DeltaNet—stem from different parameterizations of $\mA$. While this perspective offers a unifying and intuitive framework that captures a wide range of attention mechanisms, it comes with an important limitation: the state-transition terms are restricted to be scalars (in the case of Mamba-2) or identity-plus-rank-one matrices (in the case of Gated DeltaNet).

In this section, we introduce a more general framework that relaxes this scalar constraint by allowing state-transition terms (including the  thus $\lambda^{(\ell)}_t$ terms) to be matrix-valued. This extension enables richer and more expressive attention mechanisms while preserving computational efficiency.

\paragraph{Linear Attention as an SSS Tensor.}
Consider the standard linear attention mechanism with data-dependent gating and an SSS (sequentially semiseparable) mask $\mM^{\mathcal{S}}$:
\begin{equation*}
\mP = \mQ \mK^\top \odot \mM^{\mathcal{S}}, \quad \mO = \mP \mV.
\end{equation*}
In the main paper, we extend the SSS mask $\mM^{\mathcal{S}}$ to a hierarchical form $\mM^{\mathcal{H}}$. Notice that in  Mamba-2, the resulting matrix $\mP$ also inherits the same structural property, with its SSS-rank governed by the hidden dimension $d$:
\begin{equation*}
\mP_{t,s} = \mQ_t \left(\mC_{t} \cdots \mC_{s+1} \right) \mK_s^\top, \quad \text{where} \;\; \mC_{t} = \alpha_t \mI.
\end{equation*}
We now define a 4D tensor $\tM^{\mathcal{S}} \in \mathbb{R}^{\left(T \times T\right) \times \left(d \times d\right)}$ such that:
\begin{equation*}
\mP_{t,s} = \mQ_t \tM_{t,s} \mK_s^\top, \quad \text{where} \;\; \tM_{t,s} = \mC_{t} \cdots \mC_{s+1}.
\end{equation*}
Each entry $\tM_{t,s} \in \mathbb{R}^{d \times d}$ is a matrix, making $\tM^{\mathcal{S}}$ a 4D tensor. We refer to this as an SSS tensor due to its sequentially semiseparable-like structure along the temporal dimension, though this term is not yet formalized in the literature.

\begin{table}[t]
\centering
\begin{tabular}{r|ll}
\toprule
\textbf{Model} & \textbf{Temporal Structure} & \textbf{Hidden Size Structure} \\
\midrule
Mamba-2 & Semiseparable & Scaled Identity \\
Gated DeltaNet & Semiseparable & Identity plus Low-Rank \\
\emph{Log-Linear} Mamba-2 & Hierarchical & Scaled Identity \\
\emph{Log-Linear} Gated DeltaNet & Hierarchical & Identity plus Low-Rank \\
\bottomrule
\end{tabular}
\vspace{3pt}
\caption{Structural comparison of different attention variants.}
\label{table:4d-view}
\vspace{-7pt}
\end{table}

This tensor-centric view naturally accommodates  matrix-valued state transitions $\mC_t \in \mathbb{R}^{d \times d}$ with arbitrary structure, offering a richer representation than scalar- or identity-plus-rank-one-based approaches. In particular, models such as Mamba-2 and Gated DeltaNet can be interpreted as operating on 4D tensors with different hidden-dimension structures, while still preserving temporal semiseparability.\footnote{Strictly speaking, Gated DeltaNet also need to include a term $\beta_t$ from $\beta_t \vv_t \vk_t^\top$. For clarity, we omit it here, as it can be absorbed into other terms.}
\begin{equation*}
\begin{aligned}
\text{Mamba-2:} \quad \tM^{\mathcal{S}}_{t,s} = \prod_{t^\prime=t}^{s+1} \alpha_{t^\prime} \mI,
\quad\quad\quad\quad \text{Gated DeltaNet:} \quad \tM^{\mathcal{S}}_{t,s} = \prod_{t^\prime=t}^{s+1} \alpha_{t^\prime} \left(\mI - \beta_{t^\prime} \vk_{t^\prime}\vk_{t^\prime}^\top\right)
\end{aligned}
\end{equation*}
\paragraph{Log-Linear Attention as an $\mathcal{H}$ Tensor.}
We can apply our \emph{log-linear} attention to these more flexible (linear) RNNs by incorporating matrix-valued, level- and data-dependent terms $\mLambda_t^{(\ell)} \in \mathbb{R}^{d \times d}$:
\begin{equation*}
\begin{aligned}
\text{Mamba-2:} \quad \tM^{\mathcal{H}}_{t,s} = {\color{blue} \mLambda^{(\ell)}_t}\, \prod_{t^\prime=t}^{s+1} \alpha_{t^\prime} \mI,
\quad\quad\quad\quad \text{Gated DeltaNet:} \quad \tM^{\mathcal{H}}_{t,s} = {\color{blue} \mLambda^{(\ell)}_t}\, \prod_{t^\prime=t}^{s+1} \alpha_{t^\prime} \left(\mI - \beta_{t^\prime} \vk_{t^\prime}\vk_{t^\prime}^\top\right)
\end{aligned}
\end{equation*}
This formulation highlights a key insight: both Mamba-2 and Gated DeltaNet share a common semiseparable structure in the temporal dimension, but differ in how they structure the hidden dimension. Mamba-2 relies on scaled identities, while Gated DeltaNet applies identity-minus-rank-one modifications. Table~\ref{table:4d-view} summarizes these distinctions.

\section{Log-Linear Attention as $\mathcal{H}$ Matrices}
\label{appendix-subsec:variants-of-h-matrices}

We begin by introducing two classes of Hierarchical matrices ($\mathcal{H}$ matrices) following~\citet{massei2020hm}: HODLR (Hierarchically Off-Diagonal Low-Rank) matrices and HSS (Hierarchically Semi-Separable) matrices. We then show how Log-Linear Attention corresponds to a specific subclass of $\mathcal{H}$ matrices that occupies an intermediate position between these two. Finally, we discuss a further variant of $\mathcal{H}$ matrices that, in principle, allows for more refined partitioning—potentially enhancing approximation quality at the cost of increased (though constant-factor) computational complexity.

\begin{figure}[t]
	\centering
        \vspace{-20pt}
	\begin{tikzpicture}[scale=0.8] \small
	\coordinate (T18) at (0,0);  
	\coordinate (T14) at (-4,-1);
	\coordinate (T58) at (4,-1);
	\coordinate (T12) at (-6,-2);
	\coordinate (T34) at (-2,-2);
	\coordinate (T56) at (2,-2);
	\coordinate (T78) at (6,-2);
	\coordinate (T1) at (-7,-3);
	\coordinate (T2) at (-5,-3);
	\coordinate (T3) at (-3,-3);
	\coordinate (T4) at (-1,-3);
	\coordinate (T5) at (1,-3);
	\coordinate (T6) at (3,-3);
	\coordinate (T7) at (5,-3);
	\coordinate (T8) at (7,-3);
	\node (N18) at (T18) {$\mathcal{I} = \{1,2,3,4,5,6,7,8\}$};
	\node (N14) at (T14) {$\mathcal{I}_1^{(3)} = \{1,2,3,4\}$};
	\node (N58) at (T58) {$\mathcal{I}_2^{(3)} = \{5,6,7,8\}$};
	\node (N12) at (T12) {$\mathcal{I}_1^{(2)} = \{1,2\}$};
	\node (N34) at (T34) {$\mathcal{I}_2^{(2)} = \{3,4\}$};
	\node (N56) at (T56) {$\mathcal{I}_3^{(2)} = \{5,6\}$};
	\node (N78) at (T78) {$\mathcal{I}_4^{(2)} = \{7,8\}$};
	\node (N1) at (T1) {$\mathcal{I}_1^{(1)} = \{1\}$};
	\node (N2) at (T2) {$\mathcal{I}_2^{(1)} = \{2\}$};
	\node (N3) at (T3) {$\mathcal{I}_3^{(1)} = \{3\}$};
	\node (N4) at (T4) {$\mathcal{I}_4^{(1)} = \{4\}$};
	\node (N5) at (T5) {$\mathcal{I}_5^{(1)} = \{5\}$};
	\node (N6) at (T6) {$\mathcal{I}_6^{(1)} = \{6\}$};   
	\node (N7) at (T7) {$\mathcal{I}_7^{(1)} = \{7\}$};
	\node (N8) at (T8) {$\mathcal{I}_8^{(1)} = \{8\}$};
	\draw[->] (N18.south) -- (N14);
	\draw[->] (N18.south) -- (N58);  
	\draw[->] (N14.south) -- (N12);
	\draw[->] (N14.south) -- (N34);    
	\draw[->] (N58.south) -- (N56);  
	\draw[->] (N58.south) -- (N78);
	\draw[->] (N12.south) -- (N1);
	\draw[->] (N12.south) -- (N2);
	\draw[->] (N34.south) -- (N3);  
	\draw[->] (N34.south) -- (N4);  
	\draw[->] (N56.south) -- (N5);  
	\draw[->] (N56.south) -- (N6);          
	\draw[->] (N78.south) -- (N7);  
	\draw[->] (N78.south) -- (N8);          
	\end{tikzpicture}

	\begin{tikzpicture}[scale=0.3]    
	\begin{scope}
	[xshift=0cm]
	\node [above] at (4,8) {$\ell=3$};
	\draw (0,0) -- (0,8) -- (8,8) -- (8,0) -- cycle;
	\draw[pattern=north west lines, pattern color=level3] (0,0) rectangle (4,4);
	\draw[pattern=north west lines, pattern color=level3] (4,4) rectangle (8,8);
	\draw (0,4) -- (8,4);
	\draw (4,0) -- (4,8);
	\end{scope}
	\begin{scope}
	[xshift=10cm]
	\node [above] at (4,8) {$\ell=2$};
	\draw (0,0) -- (0,8) -- (8,8) -- (8,0) -- cycle;
	\draw[pattern=north west lines, pattern color=level2] (4,0) rectangle (6,2);
	\draw[pattern=north west lines, pattern color=level2] (0,4) rectangle (2,6);
	\draw[pattern=north west lines, pattern color=level2] (6,2) rectangle (8,4);
	\draw[pattern=north west lines, pattern color=level2] (2,6) rectangle (4,8);
	\draw (0,2) -- (8,2);
	\draw (0,4) -- (8,4);
	\draw (0,6) -- (8,6);
	\draw (2,0) -- (2,8);
	\draw (4,0) -- (4,8);
	\draw (6,0) -- (6,8);
	\end{scope}
	\begin{scope}
	[xshift=20cm]
	\node [above] at (4,8) {$\ell=1$};
	\draw (0,0) -- (0,8) -- (8,8) -- (8,0) -- cycle;
	\draw[pattern=north west lines, pattern color=level1] (6,0) rectangle (7,1);
	\draw[pattern=north west lines, pattern color=level1] (4,2) rectangle (5,3);
	\draw[pattern=north west lines, pattern color=level1] (2,4) rectangle (3,5);
	\draw[pattern=north west lines, pattern color=level1] (0,6) rectangle (1,7);
	\draw[pattern=north west lines, pattern color=level1] (7,1) rectangle (8,2);
	\draw[pattern=north west lines, pattern color=level1] (5,3) rectangle (6,4);
	\draw[pattern=north west lines, pattern color=level1] (3,5) rectangle (4,6);
	\draw[pattern=north west lines, pattern color=level1] (1,7) rectangle (2,8);
	
	\draw (0,1) -- (8,1);
	\draw (0,2) -- (8,2);
	\draw (0,3) -- (8,3);
	\draw (0,4) -- (8,4);
	\draw (0,5) -- (8,5);
	\draw (0,6) -- (8,6);
	\draw (0,7) -- (8,7);
	\draw (1,0) -- (1,8);
	\draw (2,0) -- (2,8);
	\draw (3,0) -- (3,8);
	\draw (4,0) -- (4,8);
	\draw (5,0) -- (5,8);
	\draw (6,0) -- (6,8);
	\draw (7,0) -- (7,8);
	\end{scope}
	\begin{scope}
	[xshift=30cm]
	\node [above] at (4,8) {$\ell=0$};
	\draw (0,0) -- (0,8) -- (8,8) -- (8,0) -- cycle;
        \foreach \i in {0,...,7} {
            \fill[level0] (\i, {7-\i}) rectangle ({\i+1}, {8-\i});
        }
	\draw (0,1) -- (8,1);
	\draw (0,2) -- (8,2);
	\draw (0,3) -- (8,3);
	\draw (0,4) -- (8,4);
	\draw (0,5) -- (8,5);
	\draw (0,6) -- (8,6);
	\draw (0,7) -- (8,7);
	\draw (1,0) -- (1,8);
	\draw (2,0) -- (2,8);
	\draw (3,0) -- (3,8);
	\draw (4,0) -- (4,8);
	\draw (5,0) -- (5,8);
	\draw (6,0) -- (6,8);
	\draw (7,0) -- (7,8);
	\end{scope}
	\end{tikzpicture}
	
	\caption{Visualization adapted from \cite{massei2020hm, kressner2019low}: This example illustrates a cluster tree of depth 3 along with the corresponding block partitions at each level. Blocks marked with stripes are stored as low-rank matrices in the HODLR format, while those filled with solid color represent dense matrices.}\label{fig:hodlr}
    \vspace{-10pt}
\end{figure}
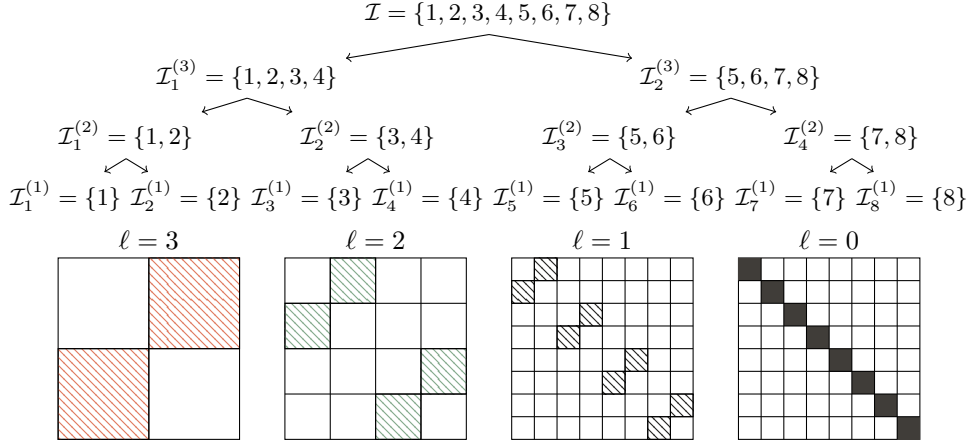
\subsection{HODLR Matrices}
\label{appendix:quasi-hierarchical}

HODLR (Hierarchically Off-Diagonal Low-Rank) matrices are structured matrices built via recursive partitioning, where off-diagonal blocks are low-rank at every level. This structure is formalized using a cluster tree~\cite{massei2020hm}.
Let $T$ be the matrix dimension, and let $\mathcal{T}$ be a perfectly balanced binary tree of depth $L$ whose nodes are subsets of $\{1, \ldots, T\}$. We say $\mathcal{T}$ is a cluster tree if: (1) the root is $\mathcal{I} = \{1, \ldots, T\}$; (2) each level partitions indices into contiguous blocks; (3) every node $\mathcal{I}^{(\ell)}i$ at level $\ell$ has two children $\mathcal{I}^{(\ell-1)}_{2i-1}$ and $\mathcal{I}^{(\ell-1)}_{2i}$ that form a disjoint partition of the parent.
See Fig.~\ref{fig:hodlr} for a visual example of such a hierarchical partitioning.

Now, let $\mM \in \mathbb{R}^{T \times T}$ be a square matrix and $\mathcal{T}$ a cluster tree as described above. We say that $\mM$ is a $(\mathcal{T}, k)$-HODLR matrix if,
\begin{equation*}
\operatorname{rank}\left(\mM[\mathcal{I}^{(\ell)}_i, \mathcal{I}^{(\ell)}_j]\right) \leq k, \quad \forall \;\;
\mathcal{I}^{(\ell)}_i, \mathcal{I}^{(\ell)}_j \in \operatorname{sibilings}\left(\mathcal{T}\right)
\end{equation*}
This hierarchical low-rank structure enables efficient $\mathcal{O}(T \log T)$ storage and matrix-vector multiplication, making HODLR matrices a core component in fast algorithms for dense matrix computations. HODLR belongs to the broader class of rank-structured matrices known as Hierarchical matrices ($\mathcal{H}$ matrices).

\subsection{HSS Matrices}
The $\mathcal{O}(T \log T)$ memory complexity of HODLR matrices arises from their recursive structure: they consist of $\mathcal{O}(\log T)$ levels, each storing low-rank factorizations that require $\mathcal{O}(T)$ space. In cases where these low-rank factors exhibit linear dependencies across levels, it is possible to exploit these relationships through nested hierarchical low-rank representations, potentially reducing the memory complexity to $\mathcal{O}(T)$ by eliminating the logarithmic factor~\cite{massei2020hm}.

Let $\mathcal{I}^{(\ell)}_i$ and $\mathcal{I}^{(\ell)}_j$ denote a pair of sibling clusters at level $\ell$ in the cluster tree $\mathcal{T}$. Define $n^{(\ell)} = 2^{\ell - 1}$ as the block size at level $\ell$. The off-diagonal block corresponding to these clusters can be parameterized as:
\begin{equation*}
\begin{aligned}
&\mM[\mathcal{I}^{(\ell)}_i, \mathcal{I}^{(\ell)}_j] = \mU^{(\ell)}_i \mSigma^{(\ell)}_{i,j} \, \left(\mV^{(\ell)}_j\right)^\top, \quad \text{where} \; &\mU^{(\ell)}_i, \mV^{(\ell)}_j \in \mathbb{R}^{n^{(\ell)} \times k}, \; \mSigma^{(\ell)}_{i,j} \in \mathbb{R}^{k \times k}
\end{aligned}
\end{equation*}
We call $\mM$ matrix a Hierarchically Semiseparable matrices (HSS) if low-rank factors at different levels are linearly related through some ``translation operators'' $\mT^{(\ell)}_{\mU}, \mT^{(\ell)}_{\mV} \in \mathbb{R}^{2k \times k}$ such that,
\begin{equation*}
\begin{aligned}
\mU_i^{(\ell)}=\left[\begin{array}{cc}
\mU_{i_1}^{(\ell-1)} & 0 \\
0 & \mU_{i_2}^{(\ell-1)}
\end{array}\right] \mT_{\mU, i}^{(\ell)}, \quad \mV_j^{(\ell)}=\left[\begin{array}{cc}
\mV_{j_1}^{(\ell-1)} & 0 \\
0 & \mV_{j_2}^{(\ell-1)}
\end{array}\right] \mT_{\mV, j}^{(\ell)}
\end{aligned}
\end{equation*}
More broadly, HSS matrices belong to a subclass of $\mathcal{H}$ matrices known as $\mathcal{H}^2$ matrices.

\subsection{Quasi-Hierarchical Matrix.}
As discussed above, when the low-rank basis matrices $\mU^{(\ell)}$ and $\mV^{(\ell)}$ exhibit linear relationships across levels $\ell$, the matrix $\mM$ reduces to a semiseparable form. In this case, both storage and matrix-vector multiplication complexities can be reduced to $\mathcal{O}(T)$. Otherwise, $\mM$ retains the general hierarchical structure with $\mathcal{O}(T \log T)$ complexity.

We define a \emph{Quasi-Hierarchical Matrix} as one in which only one of the basis sequences, either $\mU^{(\ell)}$ or $\mV^{(\ell)}$, satisfies such a linear nesting property across levels, while the other does not. The matrix $\mM^{\mathcal{H}}$ used in the Log-Linear model (Eq.~\ref{eq:h-parallel-form}) is an instance of this structure.

Both Hierarchical and Quasi-Hierarchical matrices incur $\mathcal{O}(T \log T)$ complexity for storage and computation during training. However, the use of Quasi-Hierarchical matrices plays a crucial role in enabling $\mathcal{O}(\log T)$ complexity during inference. We are not aware of a recurrent algorithm for general Hierarchical matrices that achieves logarithmic inference complexity.\footnote{In fact, our initial attempts involved using fully Hierarchical matrices, but we were unable to derive a recurrent formulation with $\mathcal{O}(\log T)$ complexity. This motivated the design of Quasi-Hierarchical matrices specifically to support efficient recurrence.}

\paragraph{Reparameterization.}
More precisely, Eq.~\ref{eq:h-parallel-form} represents a Quasi-Hierarchical matrix that has been specifically re-parameterized as a composition of the scalar weights $\lambda^{(\ell)}$ and a sequentially semiseparable (SSS) matrix $\mM^{\mathcal{S}}$. This reparameterization serves two purposes: first, to highlight the connection between our use of $\mathcal{H}$ matrices and the SSS format adopted in prior work; and second, to enable the block decomposition into a hierarchy of SSS matrices, as shown in Eq.~\ref{eq:h-decomposition}.

We present this re-parameterization below, along with its 4D tensor variant discussed in \S\ref{appendix-subsec:4d-tensor-view}, where we additionally assume that the matrices $\mU_i$ and $\mV_j$ are invertible.
\begin{equation*}
\small
\begin{aligned}
&\textbf{Matrix:} \quad &&&&\textbf{Tensor:} \\
&\mM^{\mathcal{H}}_{i, j}
\coloneqq \tau_i^{(\ell)} u_{i} v_{j} 
\Leftrightarrow \lambda_i^{(\ell)} \prod_{t=j+1}^{i} \alpha_t
\quad\quad
&&&&\tM^{\mathcal{H}}_{i, j}
\coloneqq \mT_i^{(\ell)} \mU_{i} \mV_{j}^\top 
\Leftrightarrow \mLambda_i^{(\ell)} \prod_{t=i}^{j+1} \mC_t
\\
\Rightarrow \quad
&\tau_i^{(\ell)} := \lambda_i^{(\ell)},\;
u_{i} := \prod_{t=0}^{i} \alpha_t,\;
v_{j} := \prod_{t=0}^{j} \frac{1}{\alpha_t}
&&&&\mT_i^{(\ell)} := \mLambda_i^{(\ell)},\;
\mU_{i} := \prod_{t=i}^{0} \mC_t,\;
\mV_{j}^\top := \prod_{t=0}^{j} \mC^{-1}_t
\\
\Leftarrow \quad
&\lambda_i^{(\ell)} :=\tau_i^{(\ell)} u_i v_i,\quad
a_t := \frac{r_{t-1}}{r_{t}}
&&&&\mLambda_i^{(\ell)} := \mT_i^{(\ell)} \mU_i \mV_i^\top,\quad
\mC_t := \mR_{t}^{-1} \mR_{t-1}
\end{aligned}
\end{equation*}

\begin{figure}[t]
\vspace{-15pt}
\centering
\includegraphics[width=0.99\textwidth]{figs/h-matrices-strong-and-weak-recurrent.pdf}
\caption{
\textbf{Left}: $\mathcal{H}$ matrices with strong admissibility. 
\textbf{Right}: $\mathcal{H}$ matrices with weak admissibility. 
}
\label{fig:h-matrices-strong-and-weak-recurrent}
\vspace{-3mm}
\end{figure}

\subsection{$\mathcal{H}$ Matrices with Strong and Weak Admissibility}
\label{ssec:admissability}
In the recurrent formulation of Log-Linear Attention, although there are $\mathcal{O}(\log T)$ states corresponding to different hierarchical levels, roughly half of them are zero in practice. This sparsity arises from the specific structure of HODLR matrices, which belong to a broader class of $\mathcal{H}$ matrices known as \emph{weakly admissible}~\cite{hackbusch2004hierarchical}.

Figures~\ref{fig:h-matrices-strong-and-weak} and~\ref{fig:h-matrices-strong-and-weak-recurrent} illustrate an alternative structure based on strong (or standard) admissibility. Unlike the weakly admissible variant, strongly admissible $\mathcal{H}$ matrices allow for finer-grained partitioning of the matrix, and their corresponding recurrent forms utilize all hierarchical levels.

While strong admissibility can yield more accurate approximations, it comes with a significant computational cost~\cite{hackbusch2004hierarchical}. In our early experiments, using strong admissibility in a \texttt{Triton} implementation resulted in up to a \texttt{4x} slowdown, with only marginal improvements in accuracy. As a result, we adopt the weakly admissible structure throughout this work and refer to it simply as the $\mathcal{H}$-matrix.

\begin{figure}[t]
\centering
\includegraphics[width=0.75\textwidth]{figs/h-matrices-strong-and-weak.pdf}
\caption{
\textbf{Left}: $\mathcal{H}$ matrices with strong admissibility. 
\textbf{Right}: $\mathcal{H}$ matrices with weak admissibility. 
}
\label{fig:h-matrices-strong-and-weak}
\vspace{-4mm}
\end{figure}

\section{Implementations}
\label{appendix-subsec:implementations}

\begin{lstlisting}[language=Python]
import torch
import numpy as np
import torch.nn.functional as F


def segsum(x):
    T = x.size(-1)
    x_cumsum = torch.cumsum(x, dim=-1)
    x_segsum = x_cumsum[..., :, None] - x_cumsum[..., None, :]
    mask = torch.tril(torch.ones(T, T, device=x.device, dtype=bool))
    x_segsum = x_segsum.masked_fill(~mask, -torch.inf)
    return x_segsum


def level_mask(level, T):
    if level == 0:
        return torch.eye(T, dtype=torch.bool)

    i, j = torch.meshgrid(torch.arange(T), torch.arange(T), indexing="ij")
    half = 1 << (level - 1)
    clipped = i - (i %
    valid = (i %
    return valid


def construct_H_matrix(a, L):
    T = a.size(-1)
    A = torch.exp(segsum(a))
    return sum([A * L[..., level, :].unsqueeze(-1) * level_mask(level, T) for level in range(int(np.log2(T)) + 1)])


def hattention(X, A, B, C, L, block_len=8):
    """
    Arguments:
    X: (batch, length, n_heads, d_head)
    A: (batch, length, n_heads)
    B: (batch, length, n_heads, d_state)
    C: (batch, length, n_heads, d_state)
    L: (batch, length, n_heads, num_levels) where num_levels = log2(length) + 1
    Return:
    Y: (batch, length, n_heads, d_head)
    """
    T = X.shape[1]
    assert X.dtype == A.dtype == B.dtype == C.dtype
    assert X.shape[1] %
    input_shape = X.shape
    # Rearrange into blocks/chunks
    b, cl = X.shape[0], X.shape[1]
    c = cl // block_len
    X, A, B, C, L = [x.reshape(b, c, block_len, *x.shape[2:]) for x in (X, A, B, C, L)]
    A = A.permute(0, 3, 1, 2)  # (batch, n_heads, c, block_len)
    A_cumsum = torch.cumsum(A, dim=-1)  # (batch, n_heads, c, block_len)

    num_intra_chunk_levels = int(np.log2(block_len)) + 1
    num_inter_chunk_levels = int(np.log2(T)) + 1 - num_intra_chunk_levels
    # Partition the lambda into intra-chunk and inter-chunk lambda
    L_intra, L_inter = L[..., :num_intra_chunk_levels], L[..., num_intra_chunk_levels:]
    L_intra = L_intra.permute(0, 3, 1, 4, 2)  # (batch, n_heads, num_chunks, num_levels, block_len)

    # Intra-chunk Computation
    H = construct_H_matrix(A, L_intra)  # Materialize the H matrix as a dense matrix
    Y_diag = torch.einsum("bclhn,bcshn,bhcls,bcshp->bclhp", C, B, H, X)

    # Inter-chunk Computation
    decay_states = torch.exp((A_cumsum[..., -1:] - A_cumsum))
    states = torch.einsum("bclhn,bhcl,bclhp->bchpn", B, decay_states, X)
    decay_chunk = F.pad(torch.exp(segsum(A_cumsum[..., -1])), (0, 0, 1, 0))[..., :-1, :]
    state_decay_out = torch.exp(A_cumsum)

    def compute_Y_off_level(states, level):
        mask = level_mask(level + 1, c).unsqueeze(0).unsqueeze(0)
        decay_chunk_level = decay_chunk * mask
        states = torch.einsum("bhzc,bchpn->bzhpn", decay_chunk_level, states)
        Y_off = torch.einsum(
            "bclhn,bchpn,bhcl,bclh->bclhp",
            C,
            states,
            state_decay_out,
            L_inter[..., level],
        )
        return Y_off

    Y_off = torch.zeros_like(Y_diag)
    for i in range(num_inter_chunk_levels):
        Y_off += compute_Y_off_level(states, i)

    Y = (Y_off + Y_diag).reshape(input_shape)
    return Y
\end{lstlisting}

\begin{algorithm}[H]
\caption{Chunkwise Log-Linear Attention Algorithm}
\small
\label{alg:hattention}
\begin{algorithmic}[1]
\FOR{$t \in [T / C]$}
\STATE $\mY_{[t]} = \left(\mQ_{[t]} \mK_{[t]}^\top \odot {\color{blue} \mM^{\mathcal{H}}_{[t]}} \right) \mV_{[t]}$
\ENDFOR
\STATE
\FOR{{\color{blue} $\ell \in [\log_{2} \left(T / C\right)]$}}
\FOR{$t \in [T / C]$}
\STATE $\mY_{[t]} = \mY_{[t]} + {\color{blue} \texttt{mask}^{(\ell)}_{\mQ}} \left({\color{blue} \mLambda_{[t]}^{(\ell)}} \odot \mQ_{[t]} \mS_{[t]} \right)$
\STATE $\mS_{[t+1]} = {\color{blue} \texttt{mask}^{(\ell)}_{\mA}} \left( \mA_{[t]}\mS_{[t]} \right) + {\color{blue}\texttt{mask}^{(\ell)}_{\mK}}\left( \mK_{[t]} \mV_{[t]}^\top \right)$
\ENDFOR
\ENDFOR
\RETURN $\mY$
\end{algorithmic}
\end{algorithm}

\vspace{-12pt}

A naive implementation computes each level independently using a Mamba-2-style primitive, then sums the outputs—leading to redundant memory access and kernel launches. To improve efficiency, we fuse computation across four levels into a single Triton kernel, which we found optimal given SRAM constraints on an H100.

For backpropagation, we unify gradient computation across all levels for $\nabla\mK$ and $\nabla\mV$ by analytically factoring their dependencies. This reduces kernel count and improves memory efficiency, achieving over 3× speedup compared to the naive multi-level version.

\section{Additional Experiment Details}
\label{appendix-sec:experiment-details}
For the implementation benchmarks, all experiments were conducted on an H100 GPU with a batch size of 2, using 48 attention heads, a head dimension of 64, and a chunk size of 64. In Mamba-2-style models, the attention heads are applied to $\mV$ (MVA pattern), whereas in FlashAttention-2, we adopt GQA-style attention by applying heads to $\mQ$. The dimensions of the $\mQ$ and $\mK$ states are set to 128, aligning with common training configurations.

For the MQAR experiments, we largely follow the setup described in~\citet{arora2024simple}. Models are trained and evaluated on 256-token sequences containing between 4 and 64 key-value pairs. We do not evaluate on sequences longer than those used in training (i.e., no length generalization). In (Log-Linear) Mamba-2 models, both the state and head dimensions are set to 16. For (Log-Linear) Gated DeltaNet, we use two attention heads by default, except for models with a dimension of 16, where a single head is used. We tune the learning rate and, for Log-Linear models, also tune the parameterization of $\lambda$. We run each configuration with five seeds. Training was early stopped when accuracy exceeded 99\%.

For the language modeling experiments, each run was performed on $8\times$A100 or $8\times$H100 GPUs over the course of several days. We do not tie word embeddings, use a vocabulary size of $32{,}000$, and set the initializer range to $0.006$. Training is performed with a global batch size of approximately $524$K tokens for $95$K steps (roughly $50$B tokens). We use the \texttt{flash-linear-attention} and \texttt{flame} libraries~\cite{yang2024fla,yang2025flame}, following most of their default configurations.

\begin{figure}[h]
    \centering
    \includegraphics[width=0.6\linewidth]{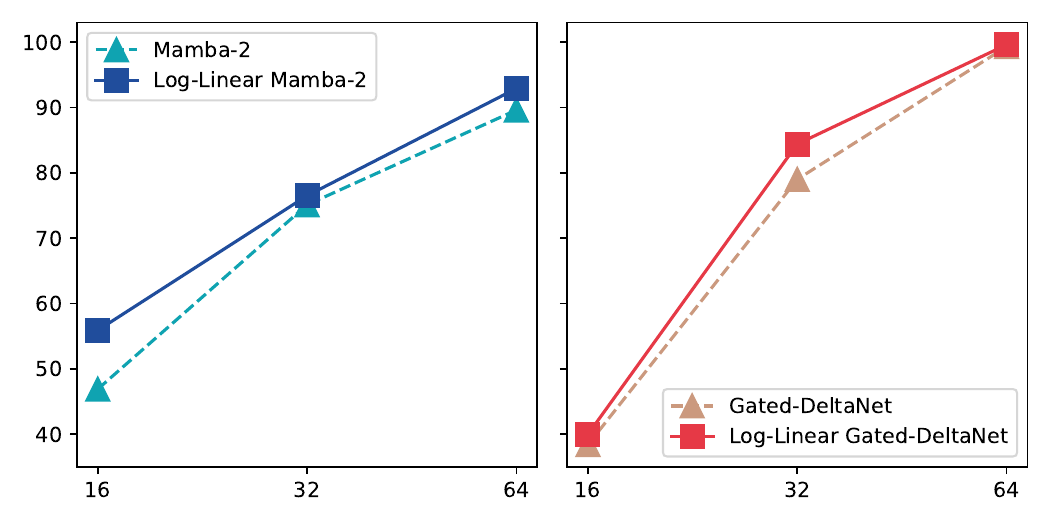}
    \vspace{-3mm}
    \caption{MQAR experiments with early stopping at 99\% accuracy.}
    \vspace{-3mm}
    \label{fig:mqar}
\end{figure}

\begin{figure}
\vspace{-20pt}
\centering
\includegraphics[width=0.99\linewidth]{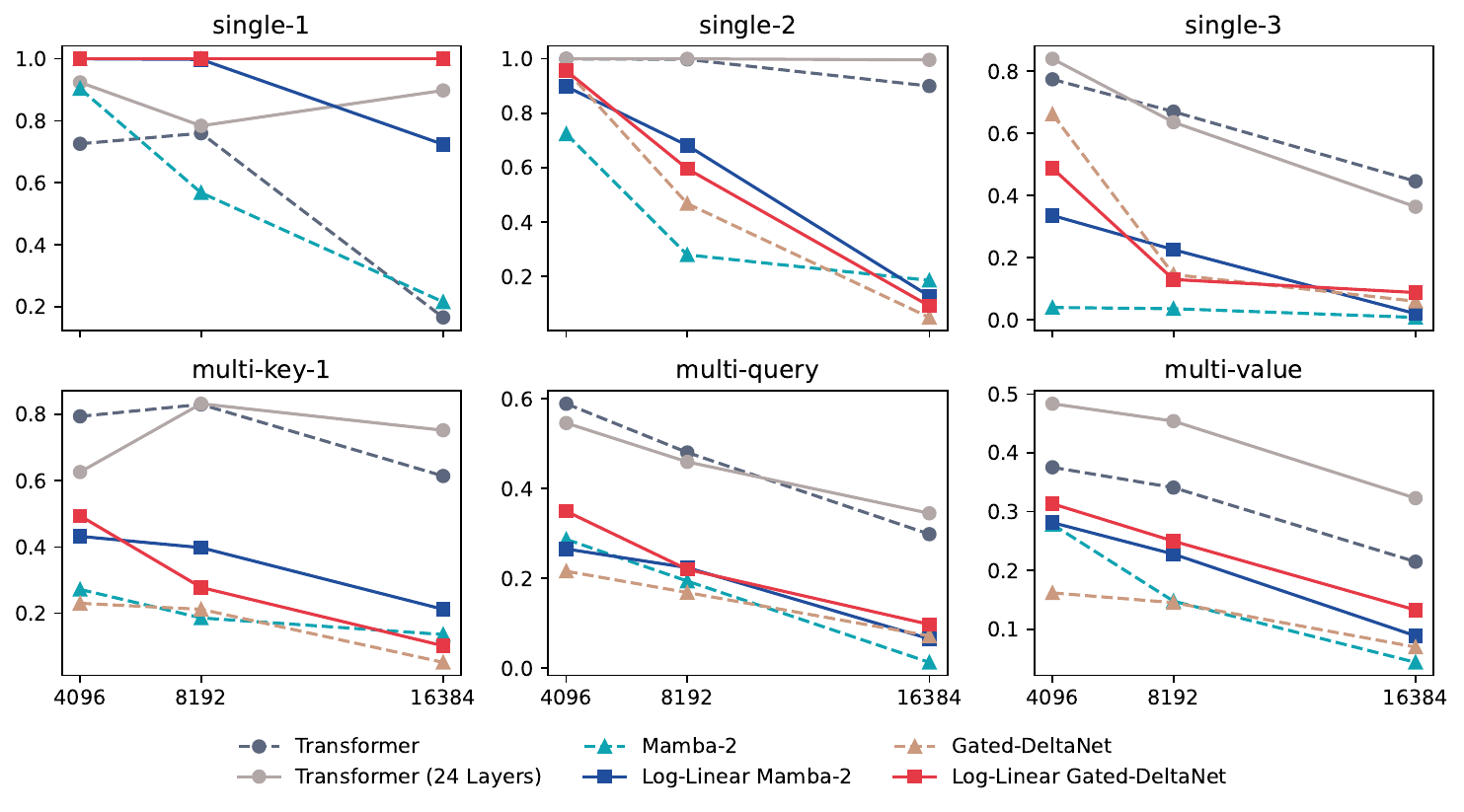}
\vspace{-2mm}
\captionof{figure}{Needle-In-A-Haystack experiments. See Table~\ref{tab:ruler} for details.}
\label{fig:ruler}
 \vspace{-3mm}
\end{figure}

\textbf{Detailed Experimental Results.}
Figures~\ref{fig:mqar} and~\ref{fig:ruler} and Tables~\ref{tab:lmeval} and~\ref{tab:based} provide detailed results.

\begin{table*}[h]
\centering
\small
\begin{tabular}{r|cc|ccccccc}
\toprule
\textbf{Model}  & \textbf{Wiki.}  &  \textbf{LMB.} &  \textbf{LMB.} & \textbf{PIQA} &    \textbf{Hella.} & \textbf{Wino.} & \textbf{ARC-e} &  \textbf{ARC-c} &  \textbf{Avg.}  \\
 & ppl $\downarrow$  &  ppl $\downarrow$  &  acc $\uparrow$  & acc $\uparrow$ &   acc\_n $\uparrow$  & acc $\uparrow$  & acc $\uparrow$ & acc\_n $\uparrow$ &  \\
\midrule
Transformer          & 21.56 & 22.14 & 38.8 & 65.1 & 39.6 & 50.7 & 45.6 & 24.5 & 44.0 \\
w/ \emph{24 Layers}  & 21.13 & 21.17 & 39.3 & 66.6 & 40.4 & 53.3 & 47.8 & 26.4 & 45.6 \\
Mamba-2              & 22.44 & 24.14 & 36.2 & 66.8 & 41.2 & 51.6 & 46.0 & 27.1 & 44.8 \\
\rowcolor{black!10}
w/ \emph{Log-Linear} & 22.11 & 21.86 & 37.0 & 66.6 & 41.1 & 51.7 & 45.5 & 27.4 & 44.9 \\
Gated DeltaNet       & 21.73 & 19.71 & 39.3 & 65.8 & 40.9 & 52.2 & 47.1 & 24.6 & 45.0 \\
\rowcolor{black!10}
w/ \emph{Log-Linear} & 21.44 & 18.08 & 40.5 & 66.1 & 41.4 & 53.9 & 46.9 & 24.9 & 45.6 \\
\bottomrule
\end{tabular}
\centering
\caption{Performance comparison on language modeling and zero-shot commonsense reasoning.} 
\label{tab:lmeval}
\end{table*}

\begin{table}[h]
\centering
\setlength{\tabcolsep}{3.5pt}  %
\small  %
\vspace{-1mm}
\begin{tabular}{r|cccc|cccc|cccc}
\toprule
 & \multicolumn{4}{c|}{\textbf{SWDE}} & \multicolumn{4}{c|}{\textbf{SQuAD}} & \multicolumn{4}{c}{\textbf{FDA}} \\[1pt]
\textbf{Model} & 512 & 1024 & 2048  & 16k
      & 512 & 1024 & 2048  & 16k
      & 512 & 1024 & 2048  & 16k
\\
\midrule
Transformer          & 47.3 & 44.6 & 45.2 & 45.4 & 34.0 & 34.5 & 34.5 & 34.5 & 72.2 & 70.8 & 72.9 & 72.2 \\
w/ \emph{24 Layers}  & 53.8 & 50.9 & 50.3 & 50.8 & 30.7 & 31.2 & 31.2 & 30.9 & 73.8 & 76.0 & 74.4 & 73.8 \\
Mamba-2              & 42.5 & 37.7 & 30.7 & 30.6 & 21.6 & 21.7 & 21.9 & 22.0 & 53.7 & 38.0 & 23.8 & 21.3 \\
\rowcolor{black!10}
w/ \emph{Log-Linear} & 41.9 & 35.6 & 28.4 & 28.5 & 25.8 & 25.9 & 25.9 & 26.1 & 53.0 & 37.5 & 20.5 & 16.6 \\
Gated DeltaNet       & 41.0 & 32.5 & 27.2 & 27.8 & 23.8 & 24.1 & 24.3 & 23.7 & 57.2 & 43.7 & 33.2 & 30.5 \\
\rowcolor{black!10}
w/ \emph{Log-Linear} & 46.2 & 39.4 & 35.3 & 35.1 & 25.2 & 25.2 & 25.3 & 25.3 & 64.9 & 53.5 & 39.1 & 30.5 \\
\bottomrule
\addlinespace
& \multicolumn{4}{c|}{\textbf{TriviaQA}} & \multicolumn{4}{c|}{\textbf{Drop}} & \multicolumn{3}{c}{\textbf{NQ}} \\
\textbf{Model} & 512 & 1024 & 2048  & 16k
      & 512 & 1024 & 2048  & 16k
      & 512 & 1024 & 2048  &  \\
\midrule
Transformer          & 48.5 & 49.6 & 48.5 & 48.5 & 22.8 & 22.8 & 22.5 & 22.3 & 24.5 & 24.3 & 24.6 & \\
w/ \emph{24 Layers}  & 46.9 & 47.0 & 46.8 & 46.8 & 22.7 & 22.4 & 22.7 & 23.0 & 24.0 & 24.4 & 24.5 & \\
Mamba-2              & 43.7 & 43.2 & 43.2 & 43.2 & 22.2 & 22.1 & 22.2 & 22.1 & 18.5 & 16.5 & 16.5 & \\
\rowcolor{black!10}
w/ \emph{Log-Linear} & 44.9 & 45.0 & 45.5 & 45.5 & 20.2 & 20.6 & 20.3 & 19.9 & 20.0 & 19.9 & 20.4 & \\
Gated DeltaNet       & 45.6 & 45.6 & 45.6 & 45.6 & 21.1 & 21.7 & 21.4 & 21.8 & 20.1 & 18.4 & 18.7 & \\
\rowcolor{black!10}
w/ \emph{Log-Linear} & 45.9 & 45.6 & 46.0 & 46.0 & 20.7 & 20.8 & 20.8 & 21.0 & 22.5 & 21.8 & 21.3 & \\
\bottomrule
\end{tabular}
\vspace{3pt}
\caption{Accuracy on retrieval tasks w/ input truncated to different lengths.}
\label{tab:based}
\end{table}

\begin{table}[h!]
\centering
\setlength{\tabcolsep}{3.5pt}  %
\small  %
\begin{tabular}{r|ccc|ccc|ccc|ccc|cc}
\toprule
& \multicolumn{3}{c|}{\textbf{Single-Doc QA}} & \multicolumn{3}{c|}{\textbf{Multi-Doc QA}} & \multicolumn{3}{c|}{\textbf{Summarization}} & \multicolumn{3}{c|}{\textbf{Few-shot}} & \multicolumn{2}{c}{\textbf{Code}} \\
 \textbf{Model} & {\tiny NQA} & {\tiny QQA} & {\tiny MFQ} 
          & {\tiny HQA} & {\tiny 2WM} & {\tiny Mus} 
          & {\tiny GvR} & {\tiny QMS} & {\tiny MNs}
          & {\tiny TRC} & {\tiny TQA} & {\tiny SSM} 
          & {\tiny LCC} & {\tiny RBP} \\\midrule
Transformer          & 11.7 & 9.7 & 20.8 & 22.4 & 29.8 & 6.7 & 13.1 & 9.4 & 3.2 & 27.5 & 28.0 & 16.2 & 23.7 & 29.8 \\
w/ \emph{24 Layers}  & 10.7 & 18.4 & 26.1 & 33.7 & 25.7 & 11.6 & 16.8 & 9.4 & 10.3 & 16.5 & 45.2 & 14.3 & 31.5 & 30.9 \\
Mamba-2              & 9.1 & 17.4 & 10.9 & 11.2 & 20.9 & 4.3 & 8.3 & 6.0 & 4.9 & 2.0 & 22.6 & 8.8 & 38.1 & 34.6 \\
\rowcolor{black!10}
w/ \emph{Log-Linear} & 9.8 & 9.6 & 15.4 & 11.5 & 22.0 & 5.1 & 5.4 & 11.1 & 4.5 & 16.5 & 21.6 & 14.9 & 31.2 & 30.3 \\
Gated DeltaNet       & 8.5 & 11.9 & 16.4 & 14.4 & 24.5 & 6.6 & 9.2 & 11.7 & 11.6 & 36.5 & 25.3 & 23.1 & 31.1 & 31.1 \\
\rowcolor{black!10}
w/ \emph{Log-Linear} & 9.9 & 6.1 & 17.6 & 17.7 & 25.2 & 7.5 & 5.5 & 11.9 & 1.9 & 8.0 & 41.1 & 23.2 & 28.3 & 29.6 \\
\bottomrule
\end{tabular}
\vspace{5pt}
\caption{Accuracy on LongBench tasks~\citep{bai2023longbench}: Narrative QA, QasperQA, MultiField QA, HotpotQA, 2WikiMultiQA, Musique, GovReport, QMSum, MultiNews, TREC, TriviaQA, SamSum, LCC, and RepoBench-P.}
\label{tab:long_bench}
\vspace{-15pt}
\end{table}

\section{LLM Usage}
In this work, large language models (LLMs) were used to enhance writing by improving clarity and conciseness, to identify relevant literature across and beyond the immediate domain, and to support research ideation, particularly in mathematics and coding.

\end{document}